\documentclass[lettersize,journal]{IEEEtran}
\usepackage{amsmath,amsfonts}
\usepackage{algorithmic}
\usepackage{array}
\usepackage{bbding}   
\usepackage{booktabs}
\usepackage{multirow}
\usepackage{bm}
\usepackage{tabularx} % for adjustable width columns 
\usepackage[caption=false,font=normalsize,labelfont=sf,textfont=sf]{subfig}
\usepackage{textcomp}
\usepackage{stfloats}
\usepackage{url}
\usepackage{verbatim}
\usepackage{graphicx}
\usepackage{cite}
\usepackage{color,xcolor}
\usepackage{colortbl} 
\usepackage{hyperref}
\usepackage[linesnumbered,lined,boxed,commentsnumbered,ruled]{algorithm2e}
\usepackage{pifont}
\usepackage{footnote}
\begin{document}

\title{Physics-Guided Detector for SAR Airplanes}

\author{
% IEEE Publication Technology, ~\IEEEmembership{Stuff,~IEEE,}
        Zhongling Huang,~\IEEEmembership{Member,~IEEE,}
        Long Liu,
        Shuxin Yang,
        Zhirui Wang,
        Gong Cheng,
        Junwei Han,~\IEEEmembership{Fellow,~IEEE}
        % <-this % stops a space
\thanks{Z. Huang, L. Liu, S. Yang, G. Cheng, and J. Han are with BRAIN Lab at School of Automation, Northwestern Polytechnical University, Xi'an, China. Z. Wang is with Key Laboratory of Network Information System Technology (NIST), Aerospace Information Research Institute, Chinese Academy of Sciences, Beijing 100190, China. This work was supported by the National Natural Science Foundation of China under Grant 62101459. \textit{Corresponding Author:} Z. Wang and G. Cheng (huangzhongling@nwpu.edu.cn, zhirui1990@126.com, gcheng@nwpu.edu.cn)}% <-this % stops a space
}

\markboth{Journal of \LaTeX\ Class Files,~Vol.~14, No.~8, August~2021}%
{Shell \MakeLowercase{\textit{et al.}}: A Sample Article Using IEEEtran.cls for IEEE Journals}


\maketitle

\begin{abstract}
The disperse structure distributions (discreteness) and variant scattering characteristics (variability) of SAR airplane targets lead to special challenges of object detection and recognition. The current deep learning-based detectors encounter challenges in distinguishing fine-grained SAR airplanes against complex backgrounds. To address it, we propose a novel physics-guided detector (PGD) learning paradigm for SAR airplanes that comprehensively investigate their discreteness and variability to improve the detection performance. It is a general learning paradigm that can be extended to different existing deep learning-based detectors with "backbone-neck-head" architectures. The main contributions of PGD include the physics-guided self-supervised learning, feature enhancement, and instance perception, denoted as PGSSL, PGFE, and PGIP, respectively. PGSSL aims to construct a self-supervised learning task based on a wide range of SAR airplane targets that encodes the prior knowledge of various discrete structure distributions into the embedded space. Then, PGFE enhances the multi-scale feature representation of a detector, guided by the physics-aware information learned from PGSSL. PGIP is constructed at the detection head to learn the refined and dominant scattering point of each SAR airplane instance, thus alleviating the interference from the complex background. We propose two implementations, denoted as PGD and PGD-Lite, and apply them to various existing detectors with different backbones and detection heads. The experiments demonstrate the flexibility and effectiveness of the proposed PGD, which can improve existing detectors on SAR airplane detection with fine-grained classification task (an improvement of 3.1\% mAP most), and achieve the state-of-the-art performance (90.7\% mAP) on SAR-AIRcraft-1.0 dataset. The project is open-source at \url{https://github.com/XAI4SAR/PGD}.
\end{abstract}


\section{Introduction}

\begin{figure}[!tb]
\centering
\includegraphics[width=0.5\textwidth]{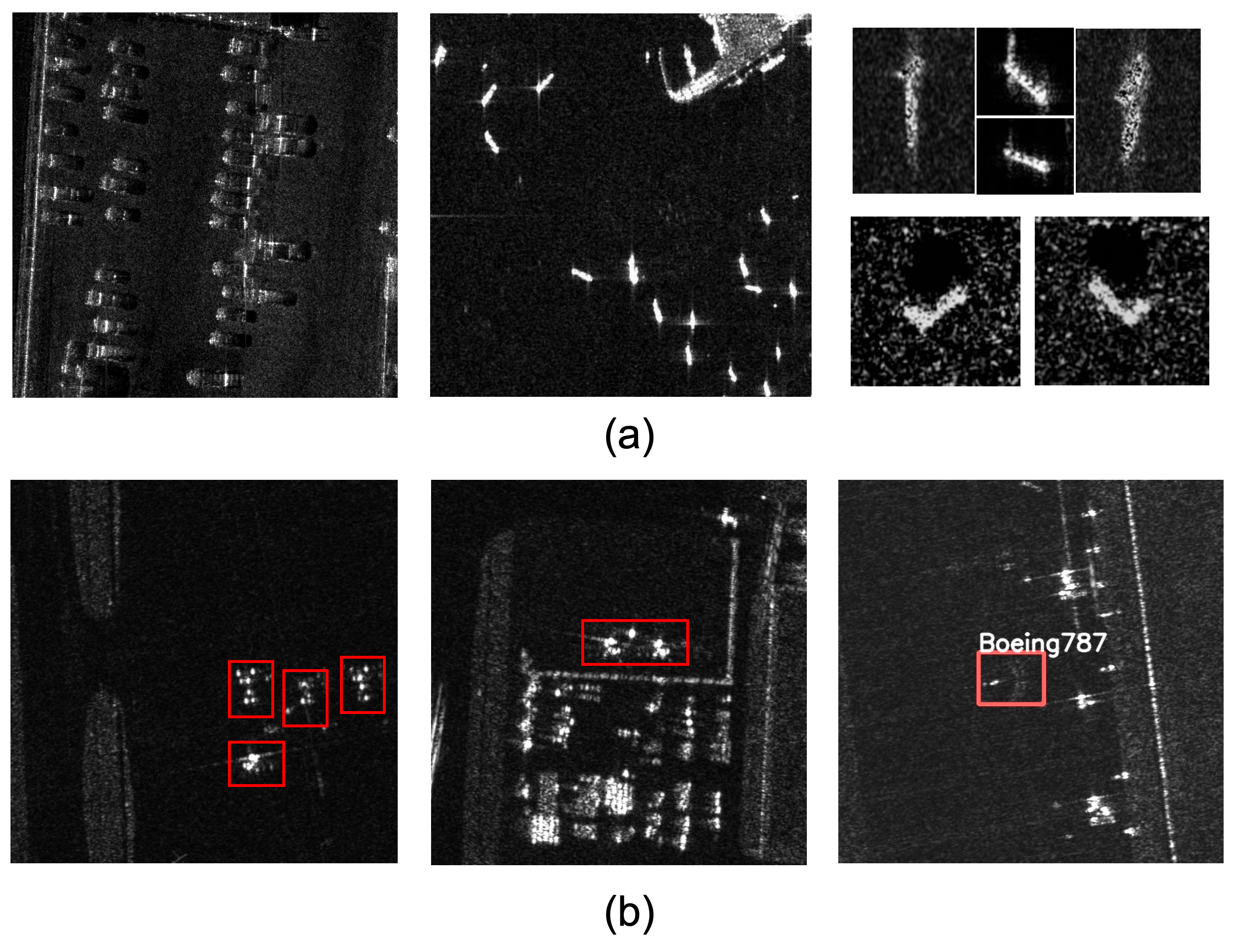}
\caption{(a) SAR ship and vehicle targets depict continuous scattering characteristics and regular shapes. (b) As a comparison, SAR airplanes are more discrete and variable.}
\label{fig:intro}
\end{figure}

SAR target detection and recognition is a fundamental task in this community, and the most concerned targets include ships, vehicles, and airplanes \cite{gao2023onboard}. Most ship and vehicle targets are with rectangle shapes, and therefore they depict relatively more stable and continuous scattering characteristics in SAR images than airplanes \cite{zhao2020pyramid,huang2024scattering}, as shown in Fig. \ref{fig:intro}. SAR airplane targets have specific characteristics, which can be summarized as discreteness and variability. The irregular geometry of airplanes introduces more complex scattering characteristics, such as edge diffraction, multiple scattering, cavity scattering, and reflection. The reflection appeared at the smooth surface of the fuselage will result in weak backscattering intensity compared with wings or engine \cite{suo2024adaptive,kang2021sfr}. In addition, the scatterings of SAR airplane targets are sparse in the complex scene, and the facilities of the airport terminals have stronger backscattering than airplanes. As a result, SAR airplane targets demonstrate more disperse structures and dramatic variations. These lead to more challenges for detection and recognition.

Data-driven deep learning-based detectors have achieved remarkable accomplishments, and they mostly require sufficient data for training\cite{gong2023edge,wang2022dense,wang2023krrnet}. Previous studies have made some progress in the field of optical remote sensing\cite{wu2024retentive,wang2024dpmnet}. Compared with optical remote sensing object detection datasets, however, SAR airplane data are limited and more versatile according to different sensors and observation parameters. It will cause severe overfitting and the complex scene at the airport terminals would significantly influence the detection result. Recently, some studies have addressed the above challenges by enhancing the scattering features or emphasizing the relations of discrete structures of SAR airplanes \cite{kang2021sfr,guo2020scattering,suo2024adaptive}. These previous methods effectively improved the detection result with better discrimination ability for SAR airplane targets and background. The fine-grained SAR airplane target recognition, on the other hand, is also challenging \cite{zhao2023classification}. The discreteness and variability of SAR airplane targets result in the inter-class similarity and intra-class diversity, which become the main obstacles for fine-grained SAR airplane recognition \cite{kang2023st,sun2022scan}. Some recent work proposed the contrastive learning to address the issue \cite{zhao2023classification}. The recent release of SAR-AIRcraft-1.0 dataset \cite{zhirui2023sar} provided a new benchmark that highlighted the main challenges of discreteness and variability for SAR airplane detection and fine-grained classification. To address the challenges in this task, the mainstream methods mostly focus on designing new model architectures and proposing optimizing methods on the basis of the current deep detectors in the field. The performance reported in the recent literature demonstrated that the performances of the existing detectors still have the room for improvement \cite{zhou2024diffdet4sar,yang2024information,li2024unleashing}.

Instead of proposing a specific detection model for SAR airplane targets, this paper aims to explore a novel general learning paradigm that can improve the existing deep learning-based detectors in perceiving the discreteness and variability of SAR airplanes. We observed that the traditional end-to-end learning paradigm has two main obstacles on SAR airplane detection task. The first lies in over-learning the complex scenes to activate them with strong responses in the feature representation while suppressing the airplane targets. The second is to under-estimate the discrete scattering details of SAR airplane instances to result in biased recognition. To this end, our motivation is to excavate the implicit prior knowledge that can provide robust and generalized representations for SAR airplanes, which guide the detector to locate the potential airplane targets more effectively in the complex scene and pay attention to the discrete scattering details for inter-class discrimination. To achieve this, we intuitively have to solving the following concerns: What implicit prior knowledge can we learn from SAR airplanes? How to leverage them to guide the detector learning? 

% 离散性，多变性，复杂性。过度学习复杂背景，使之产生较大的激活信号，影响目标本身的判别。对离散化的目标细节特征关注不足，产生理解上的语义偏差。
% \cite{,huang2024physics}

Recently, the physics-guided and injected learning paradigm for SAR image classification was proposed \cite{huang2022physically,datcu2023explainable,huang2022progress}, and has been also applied to SAR target recognition to achieve remarkable performance \cite{feng2022electromagnetic}. The main concern is to construct a self-supervised learning model for SAR images, where the pretext task is guided with the physical scattering characteristics of SAR. It can be realized with abundant SAR images without manual annotations, and the learned physics-aware features are capable of perceiving the physical properties of SAR. Inspired by them, we propose the physics-guided detector (PGD) learning paradigm for SAR airplane targets in this paper. In the previous literature, the polarimetric scattering mechanisms and attributed scattering centers were respectively considered \cite{huang2022physically,feng2022electromagnetic,huang2020hdec}. For airplane targets in single polarization amplitude SAR images, however, they are not applicable. As a result, we decide to use the discrete scattering details of SAR airplanes and their structure distribution characteristics from the amplitude images as the implicit knowledge to guide the detection model for learning a good representation. 

% The proposed PGD learning paradigm can be integrated into many deep learning-based detector architecture to enhance detection performance. 

The proposed PGD learning paradigm offers three novel components compared to existing mainstream detectors. First, we propose the Physics-Guided Self-Supervised Learning (PGSSL) to construct a surrogate task for predicting the scattering distribution of arbitrary SAR airplane targets gathered from diverse sources. The PGSSL model, once pre-trained, effectively captures the discrete scattering characteristics of various SAR airplane targets, embedded in its physics-aware features. The crucial contributing factor is to perceive the different sizes of local scattering relationships mutually within a SAR airplane target. To achieve this, we propose two different implementations based on CNN and Transformer architecture, respectively, for PGSSL. Second, the Physics-Guided Feature Enhancement (PGFE) module is developed to leverage the physics-aware features of PGSSL, enhancing the multi-scale features within the original detector. Finally, the Physics-Guided Instance Perception (PGIP) module is implemented at the detection head to learn more refined representations, facilitating fine-grained detection. We provide two implementations, denoted as PGD and PGD-Lite, respectively, emphasizing better detection results and faster inference speed. We apply them to different existing deep learning-based detectors to demonstrate the effectiveness of PGD. All experiments are conducted on SAR-AIRcraft-1.0 dataset.

The contributions are summarized as follows:

\begin{itemize}
    \item A general physics-guided detector (PGD) learning paradigm is proposed for SAR airplane detection and fine-grained classification, aiming to address the challenges of discreteness and variability. It can be extended to different deep learning-based detectors and applicable for various types of backbone models and detection heads. Several implementations, denoted as PGD and PGD-Lite, are provided to illustrate its flexibility and effectiveness.
    
    \item PGD is consist of physics-guided self-supervised learning (PGSSL), feature enhancement (PGFE) and instance perception (PGIP). PGSSL and PGFE aim to learn from airplane scattering structure distributions and further facilitate more discriminative feature representations. PGIP is designed to concern the refined scattering characteristics of each instance adaptively at the detection head. The discrete and variable scattering information of SAR airplanes are comprehensively investigated.
    
    \item Extensive experiments are conducted on SAR-AIRcraft-1.0 dataset. Based on existing detectors, we construct nine PGD models for evaluation. The results show that the proposed PGD can achieve the state-of-the-art performance on SAR-AIRcraft-1.0 dataset (90.7\% mAP) compared with other methods, and improve the existing detectors by 3.1\% mAP most. The effectiveness also proves the model-agnostic ability of PGD learning paradigm, showing significant potentials of its utility. 
\end{itemize}

The organization of this paper is summarized as follows. The literature review related to this work is introduced in Section \ref{sec:related}. Then, Section \ref{sec:method} illustrates the proposed physics-guided detector model and its implementations. The experiments and result analysis are demonstrated in Section \ref{sec:exp}. Finally, Section \ref{sec:conclusion} presents the conclusion.

\section{Related Work}
\label{sec:related}

\subsection{Object Detection in SAR Images}

In recent years, deep learning has witnessed significant achievements in object detection. This field has experienced from the initial region-based two-stage detectors (Faster R-CNN \cite{ren2015faster}, Cascade R-CNN \cite{cai2018cascade}, RepPoints \cite{yang2019reppoints}, etc), to the single-stage detectors (such as YOLO series \cite{Jocher_YOLOv5_by_Ultralytics_2020,Jocher_Ultralytics_YOLO_2023,wang2024yolov10,zhang2023superyolo} and FCOS \cite{tian2020fcos}). Recently, the research focus has shifted towards the end-to-end detection paradigm, including DINO \cite{zhang2023dino}, DETReg \cite{bar2022detreg} and DEQDet \cite{wang2023deep}. Despite the success of these methods in computer vision, the complex scattering characteristics of SAR image hinder the application prospect of them.

%  the models fail to differentiate the intricate foreground and background of SAR images and do not utilise SAR target structural information. Consequently, they cannot be directly utilised in the domain of SAR target detection.

In order to improve the suboptimal performance of existing object detection methods in the field of computer vision when applied to SAR imagery, current research efforts have been primarily directed towards three avenues of exploration. Firstly, there is an emphasis on refining the architecture of existing detectors to enhance their suitability for feature extraction from SAR images. Previous studies have shown that designing multi-scale fusion strategies for small target detection is effective\cite{guo2023save,liang2019small,zhu2024small}. Accordingly, the mainstream design include different attention mechanisms \cite{fu2021scattering,sun2022scan} and various multi-level feature fusion strategies \cite{chen2022geospatial,huang2024scattering}, so as to improve the feature representation of SAR images. Second, improving the optimization techniques are being pursued. The most commonly explored strategy is transfer learning, including effective pre-training \cite{yang2023sar,huang2019and}, domain adaptation \cite{zhou2024domain}, and meta-learning \cite{chen2022few}. These methods mainly aim to alleviate over-fitting with limited training data in SAR domain. Some other optimization strategies are also applied, such as contrastive learning to enhance the representation of the semblable fine-grained SAR targets \cite{zhao2023classification}, uncertainty estimation to improve the generalization ability of a detector \cite{huang2023uncertainty}. In addition, some recent studies propose to enrich the information for input, such as the polarization features of SAR \cite{zhao2020pyramid} and the scattering properties in spectral domain \cite{huang2024physardet}.

% One approach for SAR target detection involves incorporating attention mechanisms or context information to enhance SAR target attributes \cite{zhang2023oriented},. In recent years, some researches have added the self-attention mechanism to the SAR target detection Network, which can effectively capture the local information and the global relationship of different locations and improve the ability of the network to locate the target\cite{zhou2022pvt,han2022sean}.The other part of the work focuses on the polarization scattering information of SAR targets, and adaptively fusing the polarization characteristics to improve the scattering characteristics of SAR targets\cite{zhao2020pyramid,chen2023dpff}.
% Furthermore, other reseaches develop to boost the detection efficiency of the network on targets of different scales by designing multi-scale target feature fusion strategies.This part of the research mainly considers the semantic and texture information of feature maps of different scales, and combines features of different layers through the scale pyramid structure \cite{jiao2018densely,chen2022geospatial,fu2020anchor}.
% \textbf{However, the above methods lack the application of strong scattering characteristics of airplane targets in SAR images, which is taken into account in our proposed method.}
% SAR image quality improvement (without considering improving the quality (authenticity) of generated data)

Specifically for SAR airplane targets, the solutions generally fall within the scope of the aforementioned methodologies. However, the distinctive characteristics of SAR airplanes present unique challenges when compared to ships and vehicles. We have observed that the design of the network architecture is not consistently effective and often lacks flexibility. Additionally, the limited availability of complex-valued SAR data poses a significant obstacle to the extraction of electromagnetic attributes inherent in SAR imagery. Consequently, there is a pressing need for further in-depth research into methods and insights for addressing the detection and fine-grained classification of SAR airplanes with amplitude images in a more universally applicable manner.

% The above methods have made a great stride compared with traditional deep learning algorithms. However, they still face some problems, such as their inflexibility to multiple networks and the complexity of SAR target backgrounds. In this paper, we do not introduce a specific SAR target detection architecture. Instead, we propose a general learning paradigm specifically aimed at SAR airplane targets characterized by discrete scattering patterns. This paradigm is versatile and can be integrated into arbitrary existing detection model, allowing for implementation across various model designs.

\subsection{Prior Utilization in Object Detection}

How to introduce the prior knowledge into object detection model has attracted much attention in recent literature, due to its capability in improving the generalization in case of scarce training samples. We simply sum up as external prior guided and intrinsic prior guided methods based on the source of prior. The external prior include knowledge graph \cite{qian2024arnet}, pre-trained knowledge from external data sources \cite{cai2022bigdetection,dai2024spgc,huang2019and}, etc. The intrinsic priors encompass the existing experiences and knowledge of training samples \cite{kang2021sfr,kang2023st,dang2021msresnet,zhu2021s,lu2023improving,zhang2023object}.

Knowledge graph is a common external prior. ARNet \cite{qian2024arnet} was proposed to utilize a knowledge graph to include both common sense and expert information into the detector. Another is the knowledge learned from data itself, for example, the large-scale pre-training of vision models \cite{cai2022bigdetection}, especially the popular self-supervised learning (SSL) methods such as masked auto-encoder (MAE) \cite{he2022masked}. In remote sensing field, some foundation models are introduced and applied to downstream detection task to achieve better performance \cite{dai2024spgc,cao2023physical}. Nevertheless, it still faces challenges when transferring to SAR domain \cite{huang2019and}. A few studies used large-scale SAR images for pre-training that improved the downstream detection task to some extent \cite{yang2023sar}.

The intrinsic knowledge within training samples can also contribute to the detection task. The SAR target characteristics, for example, are considered as a strong prior in detection or fine-grained classification task. Some studies introduced the key scattering points, as well as their topology, to strengthen the feature representation \cite{kang2021sfr,kang2023st}. In addition, SSL is also applied in this case, where the labeled training images \cite{lu2023improving,zhu2021s} (sometimes including the unlabeled images \cite{dang2021msresnet}) are used, and the pre-trained weights are re-used to learn the detector \cite{dai2024spgc}. The surrogate tasks are various, such as estimating the target angle \cite{zhu2021s}, predicting the pseudo labels generated by ground truth \cite{lu2023improving}, recovering the masked area only in the region of interests \cite{zhang2023object}, etc. In these methods, however, the training data for SSL are restricted to the target task, which limit its generalization ability to some extent.

Different from the above approaches, our objective is to develop a novel self-supervised learning task that leverages the intrinsic properties of the data, alongside the utilization of externally sourced data for training purposes. To be specific, we propose a physics-guided SSL (PGSSL) to predict the scattering structure distribution of SAR airplanes, thereby reflecting their inherent characteristics. We have amassed a substantial collection of external SAR airplane target data to facilitate the learning of PGSSL model. It benefits from both external and intrinsic knowledge of SAR airplane targets, that leads to both generalization and task-oriented ability.

\section{Physics-guided Detector}
\label{sec:method}

\begin{figure*}[!htbp]
\centering
\includegraphics[width=0.7\textwidth]{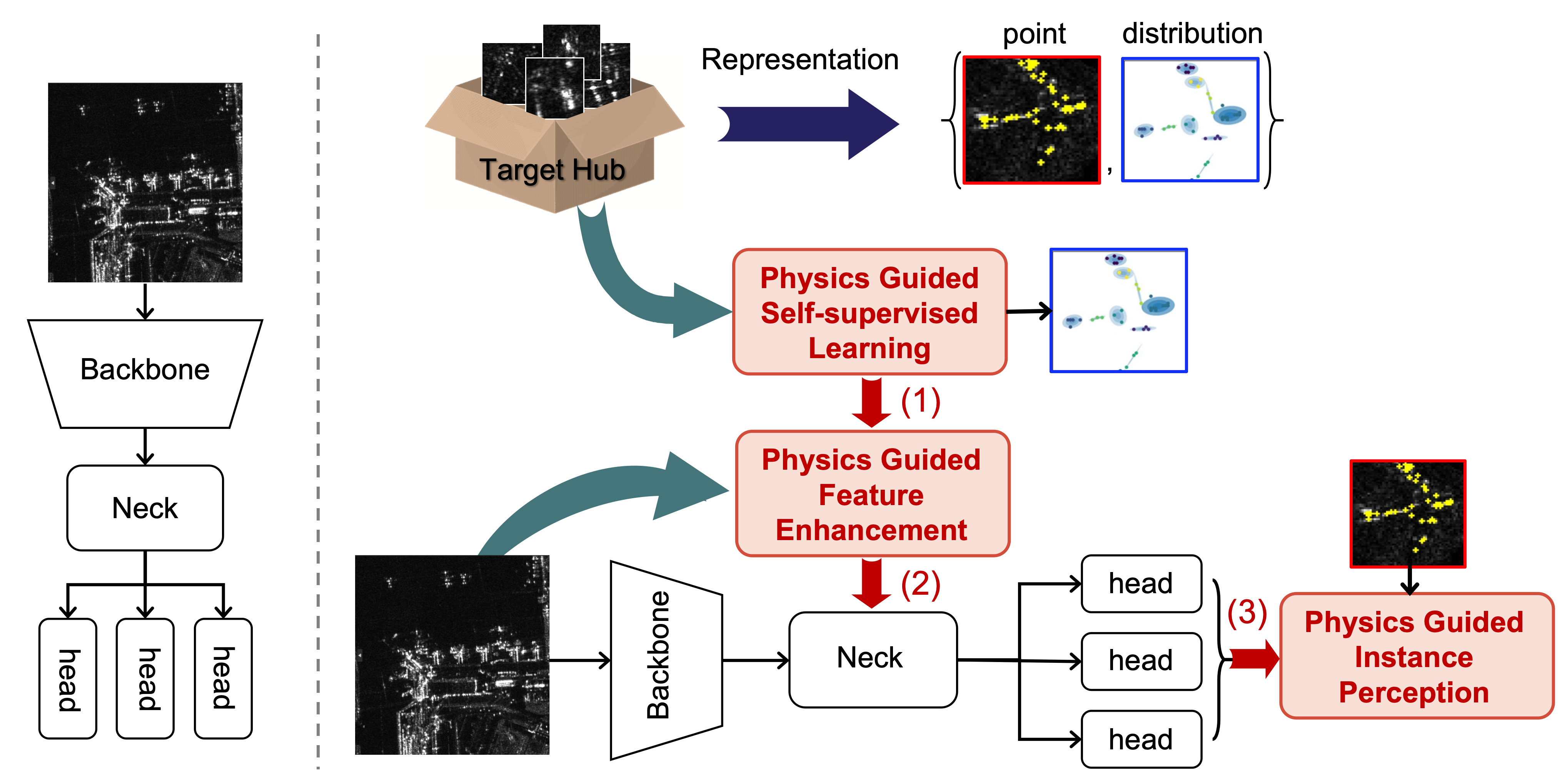}
\caption{\textbf{Left:} The general deep learning-based detector architecture. \textbf{Right: } The proposed physics-guided detector learning paradigm. It consists of (1) physics-guided self-supervised learning (PGSSL) based on a surrogate task of airplane structure distribution prediction, (2) physics-guided feature enhancement (PGFE) to improve the multi-scale features of the detector's neck, and (3) physics-guided instance perception (PGIP) at the detector's head.}
\label{fig:framework}
\end{figure*}

\subsection{Overview}

The traditional deep learning-based detectors are based on the "backbone-neck-head" architecture, as given in Fig. \ref{fig:framework} (Left). Denote the object detection training dataset as $\mathcal{D}_{\mathrm{det}}=\{x^i,y^i\}_{i=1}^{N}$, and the different parts of the detector as $f_{\mathrm{Back}}$, $f_{\mathrm{Neck}}$, and $f_{\mathrm{Head}}$. The proposed physics-guided detector (PGD) is built upon them without modifying the original architectures and can be generalized to different detectors. The overview of PGD is depicted in Fig. \ref{fig:framework} (Right), comprising three additional components: the physics-guided self-supervised learning (PGSSL), feature enhancement (PGFE), and instance perception (PGIP). The motivation is to learn the implicit prior knowledge from the discreteness and variability of SAR airplane targets to improve the representations of the general detectors. 

First, we construct a target hub containing various SAR airplane chips collected from different sources, denoted as $\mathcal{D}_{\mathrm{plane}}=\{x_{\mathrm{plane}}^i\}_{i=1}^M$. Each $x_{\mathrm{plane}}$ is represented with the collection of the discrete scattering points $S_{\mathrm{point}}$ and the estimated structure distribution $S_{\mathrm{dist}}$, which are leveraged in PGSSL and PGIP, respectively. The proposed PGSSL establishes a surrogate task of perceiving the discrete structure distributions $S_{\mathrm{dist}}$ of various SAR airplane targets $\mathcal{D}_{\mathrm{plane}}$, denoted as 
\begin{equation}
  f_{\mathrm{PGN}}:x_{\mathrm{plane}} \rightarrow S_{\mathrm{dist}} 
\end{equation}
It is notably that PGSSL is conducted on $\mathcal{D}_{\mathrm{plane}}$ and independent on the training of detectors. The details will be introduced in Section III.B.

Subsequently, PGFE is designed at the neck to enhance the physics-aware representation of the multi-scale neck features in terms of the pre-trained $f_{\mathrm{PGN}}$, denoted as
\begin{equation}
    f_{\mathrm{PGFE}} := f_{\mathrm{PGN}} \circ f_{\mathrm{Neck}}
\end{equation}
where $\circ$ will be explained with detailed implementations in Section III.C. 

At the detection head, PGIP is proposed parallel to the classification and regression branches aiming to concern the refined scattering characteristics $S_{\mathrm{point}}$ adaptively. The constraint is designed as an objective function of

\begin{equation}
    L_{\mathrm{PGIP}}(f_{\mathrm{PGIP}}(\cdot), S_{\mathrm{point}})
\end{equation}

The following sections will explain how to implement the above Equations (1), (2), and (3), respectively.

\subsection{Physics-Guided Self-supervised Learning}

\begin{figure*}[!tbp]
\centering
\includegraphics[width=1.0\textwidth]{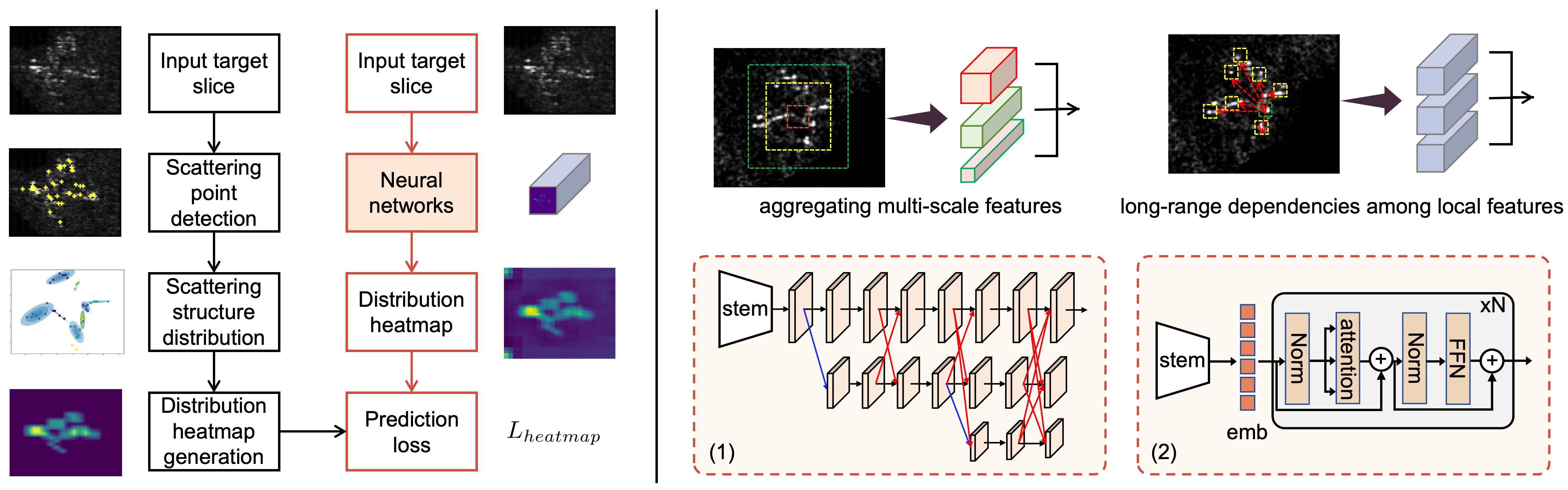}
\caption{The proposed physics-guided self-supervised learning (PGSSL) of SAR airplanes. A surrogate task of scattering structure distribution prediction is defined to construct the self-supervised learning. The proposed PGSSL can be realized with different implementations. E.g., (1) CNN based PGSSL can be designed to aggregate the multi-scale hierarchical features to represent the scattering structure distribution. (2) Transformer based PGSSL can be also designed to capture the long-range dependencies among local features of SAR airplane to represent the scattering distribution.}
\label{fig:pgssl}
\end{figure*}

The implementation of the proposed PGSSL is given in Fig. \ref{fig:pgssl}. The surrogate task is defined as predicting the scattering structure distribution $S_{\mathrm{dist}}$ for each sample $x_\mathrm{plane}$ in the collected SAR airplane targets dataset $\mathcal{D}_{\mathrm{plane}}$. We first extract the dominant scattering points of the airplane target via Harris-Laplace corner detection to obtain the collection of scattering points, denoted as $S_{\mathrm{point}}$. Then, the scattering structure distribution $S_{\mathrm{dist}}$ is estimated with a Gaussian mixture model \cite{guo2020scattering}, which can be optimized by expectation maximization (EM) method. Thus, the scattering structure distribution can be represented as:
\begin{equation} 
    p(X|\alpha,\mu,\Sigma)=\sum\limits_{k=1}^{K}\alpha_k\cdot{N(X|\bm{\mu_k},\Sigma_k)}
\end{equation}
where $K$ denotes the pre-defined number of the Gaussian distributions. $\bm{\mu_k}$ and $\Sigma_k$ represent the mean vector (i.e., the coordinate of the center) and the covariance matrix of the $k$-$\mathrm{th}$ Gaussian distribution, respectively. $\alpha_k$ is the weighted coefficient of the $k$-$\mathrm{th}$ distribution.

As shown in Fig. \ref{fig:pgssl} (Left), the distribution heatmap $S_{\mathrm{heat}}$ can be generated with the estimated statistical parameters $\bm{\mu_k}$, $\Sigma_k$, and $\alpha_k$. We first sort the K distributions according to the weight coefficients $\alpha_k$ in order to determine the importance for each scattering cluster. For each sorted Gaussian distribution, the heatmap $S_{\mathrm{heat}}^k$ is generated with $\mathcal{N}(\bm{\mu_k},\Sigma_k)$, where the regions outside the three-sigma are set to 0. Thus, the final prediction of the surrogate task is the $\mathit{K}$-channel heatmap $S_{\mathrm{heat}}$, for each channel the heatmap values are between 0 and 1. It is notably that some SAR airplane targets would have limited scattering points in the Gaussian distribution, resulting the estimated parameters unreliable to describe the true statistics information. In this case, the generated heatmap based on the estimated statistic parameters becomes unrealistic. Consequently, we consider the cluster with limited scattering points (e.g., less than 4) as a singular distribution and define a standard normal distribution to represent it. The position of the dominate scattering point in the singular distribution is set as the distribution center, and the corresponding covariance matrix $\Sigma_0$ is set to a relatively small value, i.e., $\left(
\begin{smallmatrix}
2 & 0 \\
0 & 2
\end{smallmatrix}
\right)$. We demonstrate that the threshold of the singular distribution and its variance value are two insensitive hyperparameters in this method, and we empirically set them in the experiments. To summarize, the structure distribution heatmap group $S_{\mathrm{heat}}$ is obtained from:

\begin{equation}
% \label{eq6}
S_{\mathrm{heat}}^k(i,j)=\left\{
\begin{aligned}
 e^{-\frac{1}{2}([i,j] - \bm{\mu_{k}})^{T}\Sigma_{k}^{-1}([i,j] - \bm{\mu_{k}})},& & Cnt^k \geq 4, \\
e^{-\frac{1}{2}([i,j] - \bm{\mu_0})^{T}\Sigma_{0}^{-1}([i,j] - \bm{\mu_{0}})},& & Cnt^k < 4.
\end{aligned}
\right.
\end{equation}
where $[i,j]$ denotes the position of the heatmap. $Cnt^k$ denotes the counted number of the scattering points in the $k$-$\mathrm{th}$ distribution. $\bm{\mu_0}$ and $\Sigma_0$ are the defined mean and covariance matrix of the singular distribution, where $\bm{\mu_0}$ equals to the position of the strongest scattering point in the singular distribution and $\Sigma_0=\left(
\begin{smallmatrix}
2 & 0 \\
0 & 2
\end{smallmatrix}
\right)$.

To realize Equation (1), a neural network is required to extract features from the input SAR airplane targets and predict the scattering distribution map $S_{\mathrm{heat}}$. The required deep neural network should be capable of perceiving the high-resolution features and the dependencies among the disperse components of SAR airplanes to learn the discreteness and variability successfully. Thus, we propose two implementations of $f_{\mathrm{PGN}}$ based on convolutional neural networks and vision Transformer, respectively, as given in Fig. \ref{fig:pgssl} (Right). 

\begin{figure*}[bp]
\centering
\includegraphics[width=0.9\textwidth]{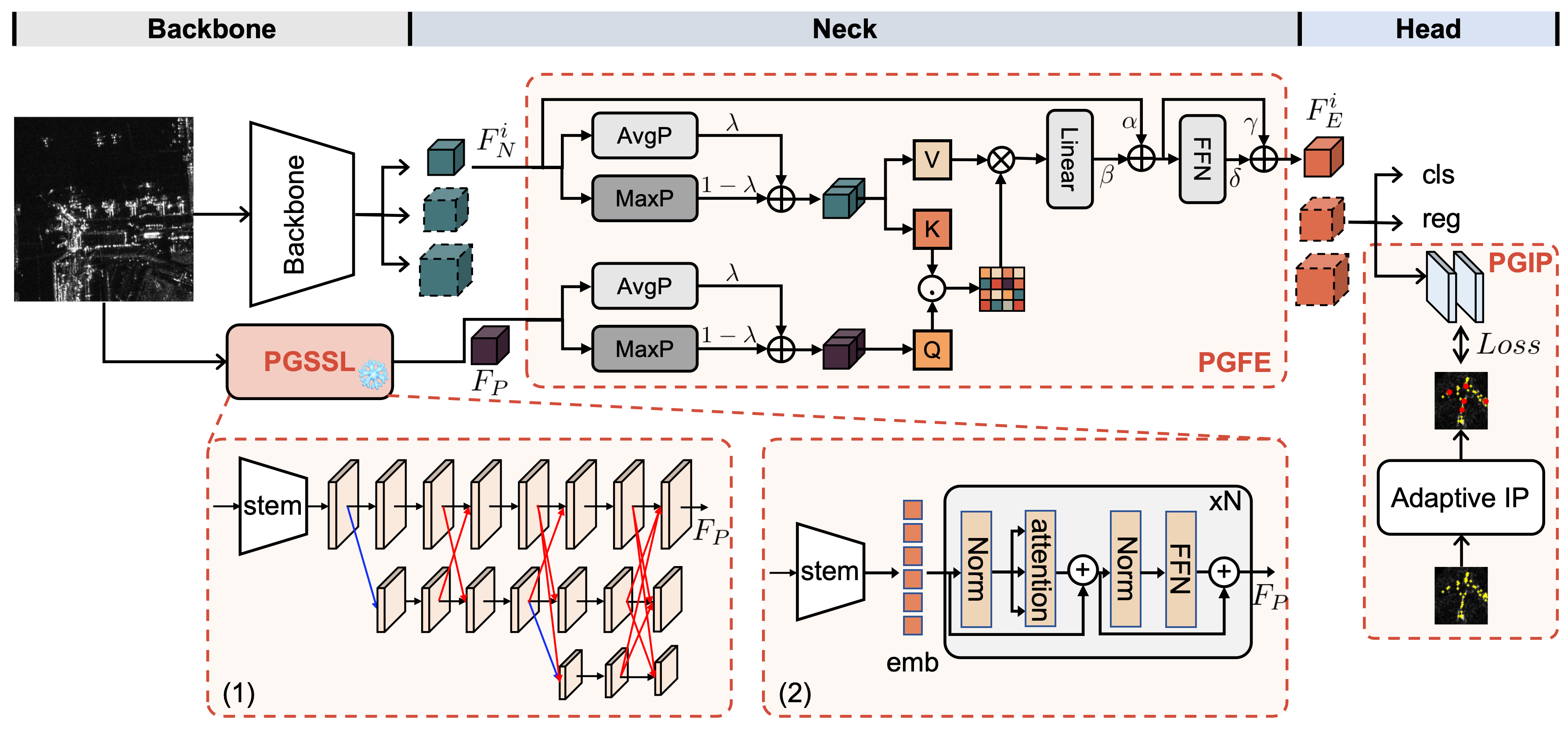}
\caption{The detailed implementation of the proposed physics-guided detector for SAR airplanes. With the input of $x$, the pre-trained PGSSL model is frozen to extract the physics-aware features $F_\mathrm{P}$ that would have representative activations in $x$.}
\label{fig:pgdet}
\end{figure*}

The first is based on HRNet \cite{sun2019deep} which maintains the  high-resolution information throughout the processing by paralleling feature maps of different resolutions and facilitating interaction between these multi-scale feature maps. It alleviates the lost of spatial structural information of the targets and the aggregated multi-scale features demonstrate better representations of the discrete components of airplanes. The other implementation is based on a Transformer architecture TransPose \cite{yang2021transpose}. The Transformer-based architecture has the advantage of modeling the long-range dependencies in images, where the multi-head attention mechanism allows the model to build the local feature interactions across the entire image. For discrete SAR airplane targets, constructing the long-range dependencies among local features of different target components is important for comprehensively representing the scattering distribution characteristics. With the input SAR airplane target $x_{\mathrm{plane}}$, the CNN-based multi-scale aggregated features or the Transformer-based long-range dependent features can be obtained as $F_\mathrm{P} = f_{\mathrm{PGN}}(x_{\mathrm{plane}})$. In order to further predict the scattering distribution heatmap $S_{\mathrm{heat}}$, a convolution layer with 1$\times$1 kernel size and $K$ channels is applied, denoted as 
\begin{equation}
    \hat{S}_{\mathrm{heat}} = conv(F_\mathrm{P}).
\end{equation}
The objective function of PGSSL is defined as the mean square error between $S_{\mathrm{heat}}$ and $\hat{S}_{\mathrm{heat}}$:
\begin{equation}
    L_{\mathrm{PGSSL}} = \frac{1}{K}\sum_k \mathrm{MSE}(S_{\mathrm{heat}}^k,\hat{S}_{\mathrm{heat}}^k).
\end{equation}

\subsection{Physics-Guided Feature Enhancement}

Subsequently, the physics-guided detector can be constructed as illustrated in Fig. \ref{fig:pgdet}. PGSSL provides a pre-trained neural network to accurately reflect the scattering distribution features of SAR airplanes. This network will exhibit substantial activation for similar discrete structures in the images. This enables the detector to be equipped with robust priors that enable the identification of potential airplane locations. At the airport scene, the terminal facilities often exhibit more pronounced scattering properties in the SAR images compared to airplane targets, and they also showcase significant activation in the feature maps. Nevertheless, they typically exhibit continuous and striped visual patterns that differ from those of airplanes. To this end, we froze the parameters in the PGSSL model and utilize the physics-aware features $F_\mathrm{P}$ to guide the detector with more meaningful representations for the discreteness properties.

The traditional detector design almost contains some feature pyramid network (FPN) style architecture to provide multi-scale features for detection head, as known as neck part. We denote them as $\{F_\mathrm{N}^i\}$. The proposed physics-guided feature enhancement (PGFE) aims to design a versatile module that can be equipped to arbitrary detector architectures with a FPN-style structure. As demonstrated in Fig. \ref{fig:pgdet}, the design of PGFE follows the cross-attention mechanism as demonstrated in \cite{shen2024icafusion}. The physics-aware features $F_\mathrm{P}$ are regarded as the query signal to calculate the weighted coefficient to enhance the multi-scale features $F_\mathrm{N}^i$ as value and key. Specifically, $F_\mathrm{P}$ are first re-sampled to have the same dimension of each $F_\mathrm{N}^i$. Then, $F_\mathrm{N}^i$ and $F_\mathrm{P}$ are compressed in advance by a parallel structure with average pooling and max-pooling to obtain query (Q), key (K), and value (V), respectively, where the trade-off parameter $\lambda$ is learnable during training. In the following, the cross-attention, linear mapping, and feed-forward processing are successively conducted to obtain the output of PGFE, denoted as $F_\mathrm{E}^i$. To summarize, PGFE can be written as:
\begin{equation}
\begin{aligned}
    &V, K = \lambda \cdot \mathrm{AvgP}(F_\mathrm{N}^i) + (1-\lambda) \cdot \mathrm{MaxP}(F_\mathrm{N}^i), \\
% \end{equation}
% \begin{equation}
    &F_\mathrm{P}^i = \mathrm{ReSample}(F_\mathrm{P}), \\
% \end{equation}
% \begin{equation}
    &Q = \lambda \cdot \mathrm{AvgP}(F_\mathrm{P}^i) + (1-\lambda) \cdot \mathrm{MaxP}(F_\mathrm{P}^i),\\
% \end{equation}
% \begin{equation}
    &Z_{\mathrm{att}}=\mathrm{Softmax}\left ( \frac{QK^{T} }{\sqrt{d_\mathrm{K}} }  \right )  \cdot V,\\
% \end{equation}
% \begin{equation}
    &Z_{\mathrm{temp}} = \beta \cdot \mathrm{Linear}(Z_{\mathrm{att}}) + \alpha \cdot F_\mathrm{N}^i,\\
% \end{equation}
% \begin{equation}
    &F_\mathrm{E}^i = \delta \cdot \mathrm{FFN}(Z_{\mathrm{temp}}) + \gamma \cdot Z_{\mathrm{temp}},
\end{aligned}
\end{equation}
where $\lambda$, $\alpha$, $\beta$, and $\gamma$ are all learnable parameters.

\subsection{Physics-Guided Instance Perception}

PGIP is designed at the detection head acting together with classification and regression losses. It is well-known that instance segmentation can substantially improve the detection performance. However, obtaining an accurate and refined instance segmentation map for a SAR airplane with a discrete structure is challenging. Several studies have suggested using the bounding box annotation to create a preliminary semantic map. However, this approach is not appropriate for SAR airplane targets that exhibit sparsity within the bounding box. We will demonstrate the issue in the experiments. In this work, we propose an adaptive instance perception method to generate the discrete semantic supervision of SAR airplanes, that constrains the detection head to learn more discriminative and precise features.

Assume that each SAR image $x$ in the training set contains $m$ annotated airplane targets, and the scattering point set $S_{\mathrm{point}}^j$ for the $j$-$\mathrm{th}$ annotated target can be obtained. We denote the additional sub-network at the detection head as $f_{\mathrm{PGIP}}$, that takes $F_\mathrm{E}^i$ as the input and outputs the single-channel prediction map $\hat{L}_i$ with a downsampled ratio $d_i$. Simultaneously, the scattering point set $S_{\mathrm{point}}^j$ is also down-sampled with a ratio of $d_i$, denoted as $S_{\mathrm{point}}^{ij}$. The proposed adaptive instance perception aims to predict the locations of the dominant scattering points for each airplane target, while ignoring the other secondary scattering locations that would impair the main topology of SAR airplane target after downsampling. Yet, we set a threshold parameter $\eta$ to decide which scattering points should be preserved during PGIP for each SAR airplane target. The positions where the response values exceed $\eta$ times the maximum response value within $S_{\mathrm{point}}^{ij}$ are set to 1, and others are set to 0. In details, we denote the PGIP supervision at the $i$-$\mathrm{th}$ detection head as $L_i$, which is obtained from:

\begin{align} \label{equ:Li}
L_i(p,q) = 
\left\{\begin{matrix}
  1, & r(p,q) \ge \eta \max(r_{ij}) \\
  0, & others
\end{matrix}\right.
\end{align}
where $r_{ij}$ is the scattering response for the $j$-$\mathrm{th}$ target with a downsampling ratio of $d_i$. Eventually, we can use the focal loss to calculate the objective function between $L_i$ and $\hat{L}_i$, denoted as 
\begin{equation}
    L_{\mathrm{PGIP}}=\sum_i \mathrm{Focal}(L_i, \hat{L}_i).
\end{equation}

\subsection{Training Process}

The training process of PGD is summarized as follows:

\begin{enumerate}
    \item \textbf{Data preparation}: Training set $\mathcal{D}_{\mathrm{det}}$ for SAR airplane target detection. For each $x_{\mathrm{det}}^i \in \mathcal{D}_{\mathrm{det}}$ with $m$ annotated targets, obtaining the scattering point set $S_{\mathrm{point}}^{ij}$, $j=1,...,m$. Collecting SAR airplane targets as dataset $\mathcal{D}_{\mathrm{plane}}$. For each $x_{\mathrm{plane}}^i \in \mathcal{D}_{\mathrm{plane}}$, obtaining the scattering distribution heatmap $S_{\mathrm{heat}}^i$.
    
    \item \textbf{PGSSL training}: Constructing a PGSSL model $f_{\mathrm{PGN}}$ based on a CNN or Transformer architecture. Training $f_{\mathrm{PGN}}$ with $\mathcal{D}_{\mathrm{plane}}$ by optimizing Equation (7) and obtaining $F_\mathrm{P}$.
    
    \item \textbf{Detector training:} Appending $f_{\mathrm{PGFE}}$ and $f_{\mathrm{PGIP}}$ on the existing detector $f_{\mathrm{det}}$ to construct the PGD model, where $f_{\mathrm{PGN}}$ is frozen during detector training. The parameters in $f_{\mathrm{det}}$, $f_{\mathrm{PGFE}}$, and $f_{\mathrm{PGIP}}$ are trained with $\mathcal{D}_{\mathrm{det}}$ by optimizing $L_{\mathrm{cls}}+L_{\mathrm{reg}}+L_{\mathrm{PGIP}}$. The additional modules $f_{\mathrm{PGFE}}$ and $f_{\mathrm{PGIP}}$ only introduce limited learnable parameters.
\end{enumerate}

\begin{table*}[!htbp]
    \centering
    \caption{We construct nine PGD models and nine PGD-Lite models based on three backbone architectures (ResNet18, ResNet50, CSPDarkNet) and three detection heads (YOLOv5, YOLOv8, YOLOv10). The results demonstrate that the proposed PGD and PGD-Lite can improve different types of existing detectors. The improvements are marked in \textcolor{red}{RED}.}
    \begin{tabular}{ccccccccc|c}
    \toprule
Method      & Backbone & Boeing787   & A220       &A320/321     &Boeing737       & A330         & ARJ21           & other         & mAP                                                  \\
\midrule
\rowcolor{gray!20}
YOLOv5        & ResNet18  & 84.1          & 88.6         & 97.6          & 81.1      & 88.6      & 86.2         & 79.7         & 86.6                                                 \\
PGD-YOLOv5      & ResNet18  & 89.3 & 92.4 & 98.8 & 82.7 & 96.1 & 88.0 &80.5 & \textbf{89.7 \textcolor{red}{(+3.1)}} \\
PGD-YOLOv5-Lite & ResNet18  &89.5 &91.5       & 98.2                        & 82.7 & 96.1 & 87.6 & 80.3 & \textbf{89.4 \textcolor{red}{(+1.9)}}         \\
\midrule
\rowcolor{gray!20}
YOLOv5        & ResNet50  & 93.1          & 86.1         & 98.0          & 76.9      & 94.1      & 85.2         & 81.8         & 87.9                                                 \\
PGD-YOLOv5      & ResNet50  & 92.0 & 90.1 & 96.0 & 83.6 & 97.6 & 88.1 &85.1 &\textbf{90.4 \textcolor{red}{(+2.5)}} \\
PGD-YOLOv5-Lite & ResNet50  &91.0 &91.4       & 97.2                        & 82.3 & 92.0 & 90.6 & 85.5 & \textbf{90.0 \textcolor{red}{(+2.1)}}         \\ 
\midrule
\rowcolor{gray!20}
YOLOv5        & CSPDarkNet  & 89.7          & 83.1         & 94.4          & 82.0      & 91.3      & 85.9         & 85.7         & 87.5                                                 \\
PGD-YOLOv5      & CSPDarkNet  & 88.2 & 90.6 & 97.7 & 82.7 & 96.1 & 89.4 &83.1 &\textbf{89.7 \textcolor{red}{(+2.2)}} \\
PGD-YOLOv5-Lite & CSPDarkNet  &89.8 &89.4       & 98.1                        & 81.2 & 95.3 & 88.3 & 83.5 & \textbf{89.5 \textcolor{red}{(+2.0)}}         \\
\midrule
\rowcolor{gray!20}
YOLOv8       & ResNet18  & 89.9            & 87.6            & 96.9              & 79.6     & 90.2      & 89.3          & 81.9     & 88.1                                                 \\
PGD-YOLOv8      & ResNet18  &94.2       &85.9            &98.2         &78.7       &98.6             & 89.1            &82.5              &\textbf{89.6 \textcolor{red}{(+1.5)}}                  \\
PGD-YOLOv8-Lite & ResNet18  & 90.6         & 87.4             & 98.5      & 82.0         & 96.6            & 87.1   & 79.6        & \textbf{88.8 \textcolor{red}{(+0.7)}}                                           \\
\midrule
\rowcolor{gray!20}
YOLOv8       & ResNet50  & 91.2            & 85.3            & 97.5              & 80.7     & 95.0      & 87.4          & 84.1     & 88.7                                                 \\
PGD-YOLOv8      & ResNet50  &93.1       &90.0            &99.0         &81.9       &96.4             & 88.4            &86.3              &\textbf{90.7 \textcolor{red}{(+2.0)}}                  \\
PGD-YOLOv8-Lite & ResNet50  & 92.8         &89.9              & 97.1      &80.0          & 95.8            &88.0    & 83.4        &\textbf{89.6 \textcolor{red}{(+0.9)}}                                           \\
\midrule
\rowcolor{gray!20}
YOLOv8       & CSPDarkNet  & 91.1            & 85.6            & 97.2              & 74.7     & 93.1      & 86.1          & 83.1     & 88.4                                                 \\
PGD-YOLOv8      & CSPDarkNet  &92.2       &86.8            &98.6         &84.9       &91.9             & 87.8            &87.2              &\textbf{89.9 \textcolor{red}{(+1.5)}}                  \\
PGD-YOLOv8-Lite & CSPDarkNet  & 90.5         & 85.8             & 98.3      & 80.4         & 97.0            & 88.7   & 82.7        & \textbf{89.0 \textcolor{red}{(+0.6)}}                                           \\
\midrule
\rowcolor{gray!20}
YOLOv10   & ResNet18    & 90.4                  & 80.5       & 97.1              & 75.5                  & 98.4          & 87.3               & 76.2            & 86.5        \\
PGD-YOLOv10          & ResNet18  &92.2       &83.6       &98.4           &76.1      &97.3        & 86.3      & 79.1        & \textbf{87.6 \textcolor{red}{(+1.1)}}                                                    \\
PGD-YOLOv10-Lite     & ResNet18  & 91.5        & 81.2      & 98.7     & 73.9         & 98.5      & 84.7         & 80.2         & \textbf{87.0 \textcolor{red}{(+0.5)}}                                          \\
\midrule
\rowcolor{gray!20}
YOLOv10   & ResNet50    & 88.4                  & 84.3       & 98.1              & 75.0                  & 93.7          & 86.2               & 81.3            & 86.7        \\
PGD-YOLOv10          & ResNet50  & 89.6     &87.9       & 98.6          &74.2      & 97.5       & 85.4      & 80.5        & \textbf{87.7 \textcolor{red}{(+1.0)}}                                                     \\
PGD-YOLOv10-Lite     & ResNet50  &  89.3       &82.1       & 97.5     &75.5          & 98.7      &  85.6        &  81.3        &\textbf{87.1 \textcolor{red}{(+0.4)}} \\
\midrule
\rowcolor{gray!20}
YOLOv10   & CSPDarkNet    & 90.2                  & 80.6       & 97.5              & 75.3                  & 93.2          & 81.0               & 78.4            & 85.2        \\
PGD-YOLOv10          & CSPDarkNet  &90.5       &84.5       &98.7           &76.6      &92.0        & 87.0      & 79.8        & \textbf{87.0 \textcolor{red}{(+1.8)}}                                                     \\
PGD-YOLOv10-Lite     & CSPDarkNet  & 92.8        & 83.7      & 98.4     & 75.7         & 93.2      & 85.6         & 78.9         & \textbf{86.9 \textcolor{red}{(+1.7)}} \\\bottomrule
    \end{tabular}
    \label{tab:PGD}
\end{table*}

\section{Experiments}
\label{sec:exp}
\subsection{Dataset and Settings}

We adopt SAR-AIRcraft-1.0 dataset \cite{zhirui2023sar} as $\mathcal{D}_{\mathrm{det}}$ for evaluation. The data were obtained from Gaofen-3 that contain 4,368 images and 16,463 airplane instances, covering seven airplane categories, namely A220, A320/321, A330, ARJ21, Boeing737, Boeing787, and other. There are 3,047 images in the training set, 442 images for validation and 879 images for testing. The image size varies from 800$\times$800 to 1500$\times$1500. In our experiment, the training and testing stages used an identical adjustment of the image size to 1024$\times$1024. At PGSSL stage, we collect different SAR airplane target chips from various data sources to form $\mathcal{D}_{\mathrm{plane}}$, including SADD \cite{zhang2022sefepnet}, TerraSAR-X \cite{pan2016airplane}, Multiangle SAR Dataset \cite{ruyi2022multiangle}, and SAR-ACD \cite{saracd}. Data augmentation is applied with Gaussian noise, translation, mirroring, rotation, etc, to obtain 10,240 samples in total. The training, validation, and testing sets are split with 7,300, 1,970, and 970, respectively. The input image size is set to 256$\times$192 at PGSSL stage.

The PGSSL applies stochastic gradient descent (SGD) optimizer with a momentum of 0.9. The initial learning rate is set to 0.001, and it decreases to 10$^{-4}$ and 10$^{-5}$ at the 100$^{th}$ and 150$^{th}$ epoch, respectively. The total training epoch is set to 170. The training of the detector was optimized by SGD with an initial learning rate of 0.005, a momentum of 0.937, and weight decay of 0.0005. The linear Warm-up schedule is utilized to set the learning rate, where the warm-up epoch is 3. The warmup initial momentum is 0.8. The warmup initial bias learning is 0.1. The batchsize is set to 4. The detector is trained for 300 epochs in total. In the ablation studies, the detector is trained for 50 epochs in total and other parameters are consistent with above settings, considering the convergence speed and calculation time of the models. We conducted all the experiment on a server equipped with four RTX 3090 GPUs while using a single RTX 3090 GPU for inference.

\subsection{Effectiveness of PGD}

We first evaluate the effectiveness of the proposed PGD by constructing different models based on existing detectors. As demonstrated in Table \ref{tab:PGD}, we experimented three backbone architectures (ResNet18, ResNet50, and CSPDarkNet) and three detection heads of YOLOv5 \cite{Jocher_YOLOv5_by_Ultralytics_2020}, YOLOv8 \cite{Jocher_Ultralytics_YOLO_2023}, and YOLOv10 \cite{wang2024yolov10}. The detection heads of YOLOv5, YOLOv8, and YOLOv10 are respectively anchor-based, anchor-free, and anchor-free without NMS. For each model, we implement the PGD and PGD-Lite version of it, demonstrating using CNN (HRNet) or Transformer (TransPose) to realize PGSSL respectively.

It can be observed that the proposed PGD model improves the mAP of 1.1\%, 1.5\%, and 3.1\% for YOLOv10, YOLOv8, and YOLOv5 on ResNet18, respectively. The mAP of the model has increased by 1.0\%, 2.0\% and 2.5\% respectively on ResNet50. The model's mAP improved by 1.8\%, 1.5\% and 2.2\% on CSPDarkNet, respectively. Among them, PGD-YOLOv5 (ResNet18) achieves the most significant improvement of 3.1\% mAP, and PGD-YOLOv8 (ResNet50) achieves the best mAP of 90.7\%. PGD-Lite denotes the implementation of applying the TransPose architecture for PGSSL with fewer parameters and faster speeds than PGD. PGD-YOLOv5-Lite performs slightly inferior than PGD-YOLOv5, but still achieves significant progress compared with the original YOLOv5 (improving 1.9\%, 2.1\%, and 2.0\% for ResNet18, ResNet50, and CSPDarkNet, respectively). In a word, the results demonstrate the effectiveness of the proposed PGD learning paradigm integrated with various deep detectors, no matter it is anchor-based or anchor-free, proving its flexibility and model-agnostic characteristic.

\begin{table*}[htbp]
\centering
\caption{Comparisons of the proposed method with other 13 state-of-the-art (SOTA) methods on SAR-AIRcraft-1.0. The \textbf{Best} and the \underline{Second-Best} results are marked in bold and underline, respectively.}
\label{tab:result}
\begin{tabular}{ccccccccc|c}
\toprule
Method                                    & Backbone       & Boeing787     & A220       & A320/321    & Boeing737   & A330      & ARJ21       & other         & mAP                  \\ \midrule
\cellcolor{gray!25}\textit{Query-based Detectors} &               &                &            &            &              &           &           &                 &                \\
\S DINO\cite{zhang2023dino}                                      & ResNet50               & \underline{92.8}           & 77.9       & 95.9       & 74.3         & 81.7      & 87.6      & \underline{86.2}        & 85.2    \\
\S DETReg\cite{bar2022detreg}                                    &ResNet50               & \textbf{93.1}           & 73.5       & 86.0       & 69.3         & 86.8      & 80.9      & 83.6        & 87.0    \\
\S DEQDet\cite{wang2023deep}                                    &ResNet50               & 81.8           & 71.6       & 92.7       & 68.3         & 89.1      & 76.2      & 84.1 & 80.3              \\ \hline
\cellcolor{gray!25}\textit{One-stage Detectors}        &               &                &            &            &              &           &           &                                 &                         \\
\dag FCOS\cite{tian2020fcos}                            & ResNet50       & 46.8            & 60.2       & 65.6       & 41.9         & 30.8      & 57.6      & 62.6 & 55.2       \\
\S SuperYOLO\cite{zhang2023superyolo}                                &CSPDarkNet              & 91.6            & 84.8       & \underline{98.6}       & 73.8         & 94.3        & 85.6    & 83.1  & 84.1  \\
\S YOLOv5\cite{Jocher_YOLOv5_by_Ultralytics_2020}                                & ResNet50   & \textbf{93.1}            & 86.1      & 98.0        & 76.9          & 94.1      & 85.2     & 81.8  & 87.9  \\ 
\dag YOLOv8\cite{Jocher_Ultralytics_YOLO_2023}                                & ResNet50 & 91.2            & 85.3            & 97.5              & 80.7     & 95.0      & 87.4          & 84.1     & 88.7  \\
\S YOLOv10\cite{wang2024yolov10}                                   & ResNet50    & 88.4& 84.3  & 98.1              & 75.0                  & 93.7          & 86.2               & 81.3            & 86.7  \\\hline
\cellcolor{gray!25}\textit{Two-stage Detectors}         & &  &  &  &  &  &  &   &      \\
\dag Faster R-CNN\cite{ren2015faster}                                      & ResNet50     & 72.9              & 78.5       & 97.2       & 55.1         & 85.0        & 74.0    & 70.1  & 76.1  \\
\dag Cascade R-CNN\cite{cai2018cascade}                            & ResNet50       & 68.3            & 74.0      & 97.5       & 54.5           & 87.4      & 78.0     & 69.1  & 75.7  \\
\dag RepPoint\cite{yang2019reppoints}                                   & ResNet50 & 51.8              & 71.4       & 97.9       & 55.7         & 89.8        & 73.0    & 68.4  & 72.6  \\ \hline
\cellcolor{gray!25}\textit{Detectors Designed for SAR}                              & &  &  &  &  &  &  &   &      \\
\dag SKGNet\cite{fu2021scattering}                                    &ResNet50             & 69.6            & 66.4        & 78.2       & 65.1         & 79.3       & 65.0     & 71.4  & 70.7                     \\
\dag SANet\cite{zhirui2023sar}                                    &ResNet50            & 70.8                 & 80.3                 & 94.3         & 59.7        & 88.6    & 78.6     & 71.3     & 77.7     \\ \hline
\cellcolor{gray!25}\textbf{Ours}          &  &   &   &    &   &    &    &    &     \\
PGD-YOLOv5                                       & ResNet50  & 92.0 & \underline{90.1} & 96.0 & \textbf{83.6} & \textbf{97.6} & 88.1 &85.1 &\underline{90.4}                                \\ 
PGD-YOLOv5-Lite & ResNet50  &91.0 &\textbf{91.4}       & 97.2                        & \underline{82.3} & 92.0 & \textbf{90.6} & 85.5 & 90.0  \\
PGD-YOLOv8                                  & ResNet50  &\textbf{93.1}      &90.0            &\textbf{99.0}         &81.9       & \underline{96.4}             & \underline{88.4}            &\textbf{86.3}              &\textbf{90.7}          \\
PGD-YOLOv8-Lite & ResNet50  & \underline{92.8}         &89.9              & 97.1      &80.0          & 95.8            &88.0    & 83.4        &89.6  \\
% PGD-YOLOv10          & ResNet50  &89.6      &87.9       &\underline{98.6}           &74.2      & 97.5       & 85.4      & 80.5        & 87.7                     \\
% PGD-YOLOv10-Lite     & ResNet50  &  89.3       &82.1       & 97.5     &75.5          & \textbf{98.7}      &  85.6        &  81.3        &87.1  \\
\bottomrule
\multicolumn{4}{l}{\small \dag \; are the result from \cite{li2024unleashing}.}\\
\multicolumn{4}{l}{\small \S \; are implemented by us.}\\
\end{tabular}
\end{table*}

\subsection{Comparison with State-of-the-arts}

\begin{figure*}[htb]
    \begin{minipage}[b]{0.19\linewidth}
      \centering
      \includegraphics[width=0.9\linewidth]{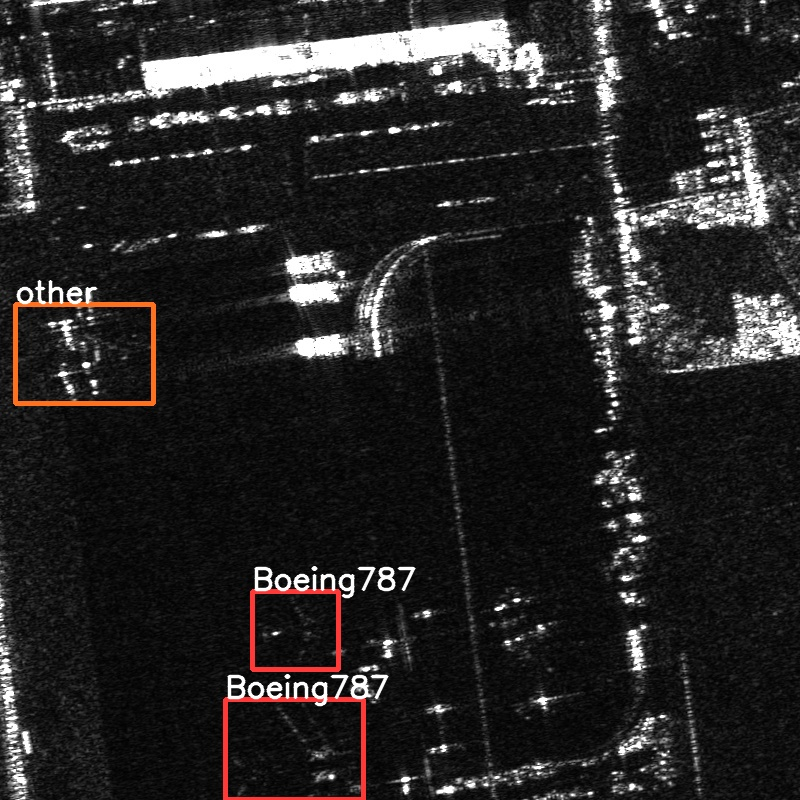}\vskip+3pt
      \includegraphics[width=0.9\linewidth]{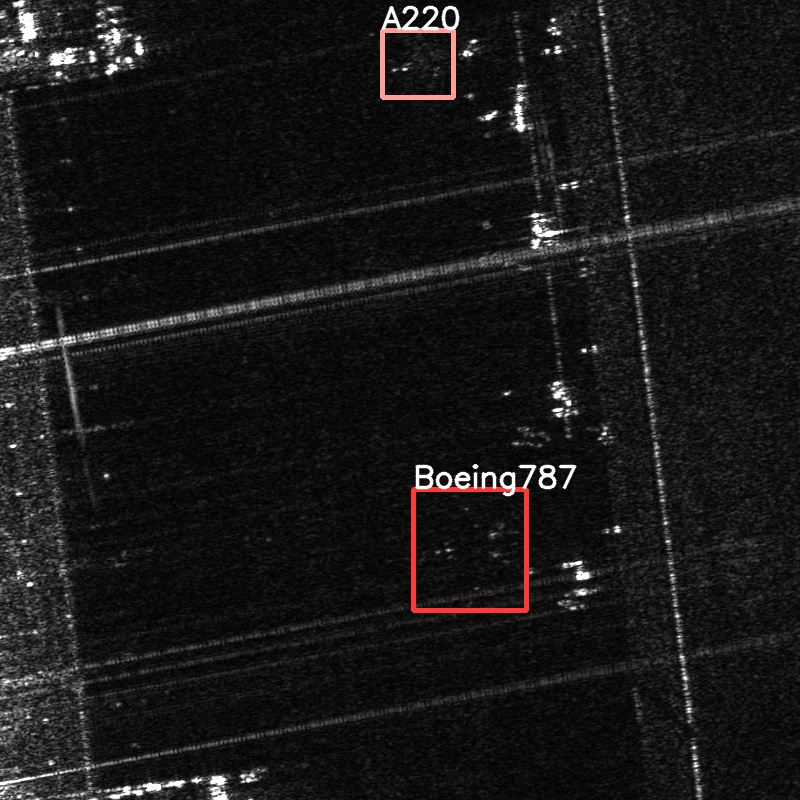}\vskip+3pt
      \includegraphics[width=0.9\linewidth]{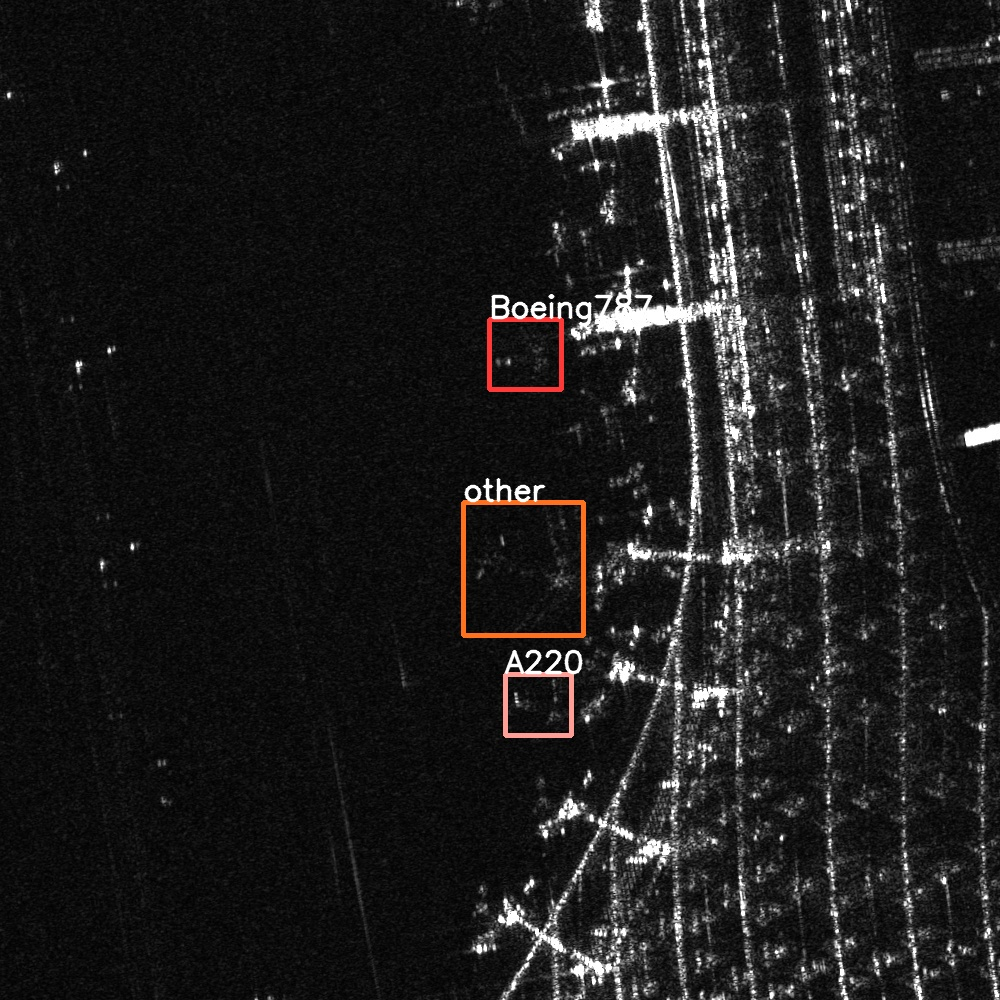}\vskip+3pt
      \includegraphics[width=0.9\linewidth]{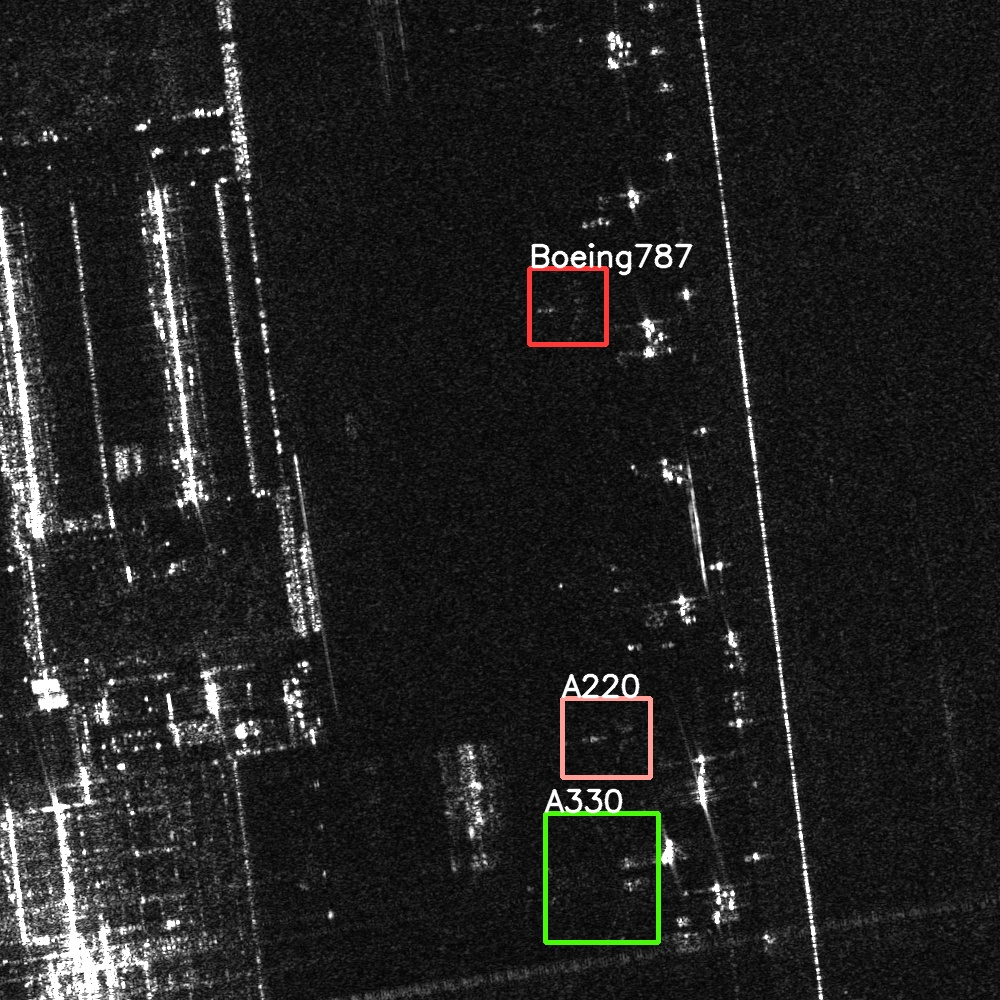}
      \centerline{(a)GT}\medskip
    \end{minipage}
    \begin{minipage}[b]{0.19\linewidth} 
        \centering
         \includegraphics[width=0.9\linewidth]{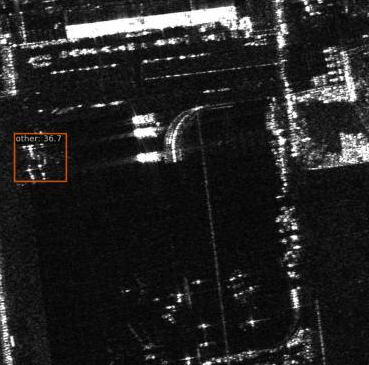}\vskip+3pt
         \includegraphics[width=0.9\linewidth]{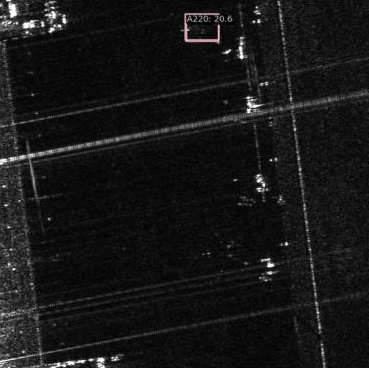}\vskip+3pt
         \includegraphics[width=0.9\linewidth]{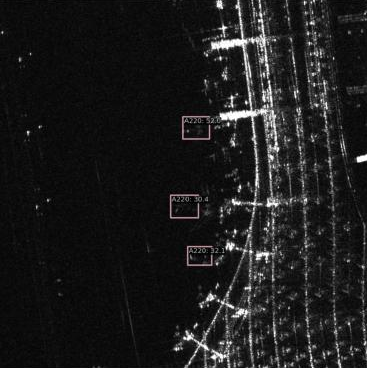}\vskip+3pt
         \includegraphics[width=0.9\linewidth]{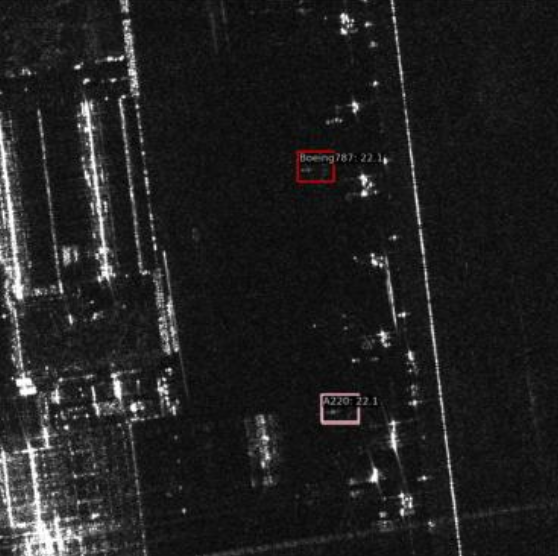}
    \centerline{(b)DEQDet.}\medskip
      \end{minipage}
    \begin{minipage}[b]{0.19\linewidth} 
        \centering
        \includegraphics[width=0.9\linewidth]{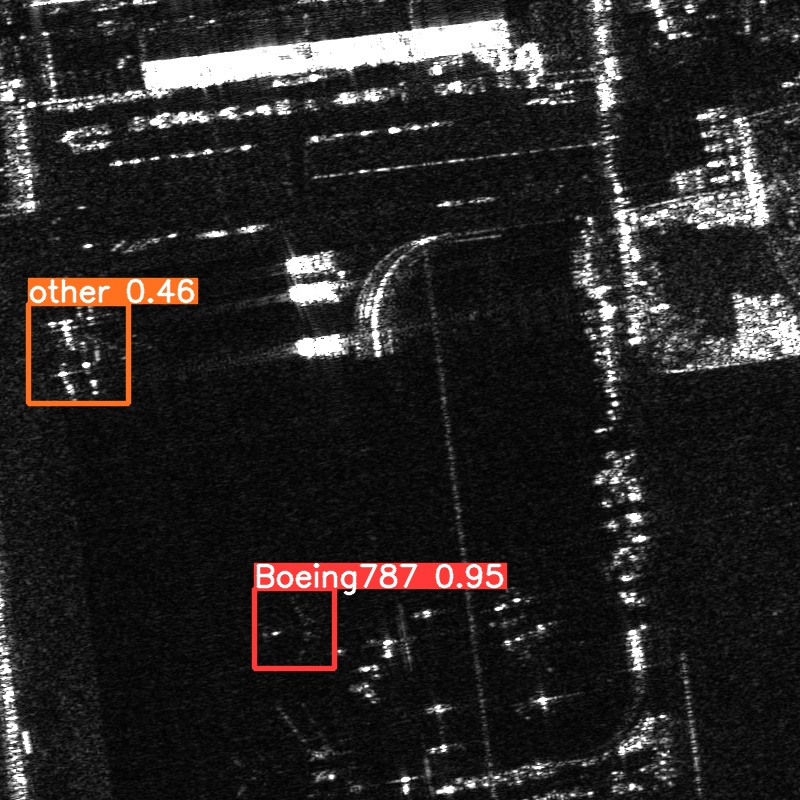}\vskip+3pt
        \includegraphics[width=0.9\linewidth]{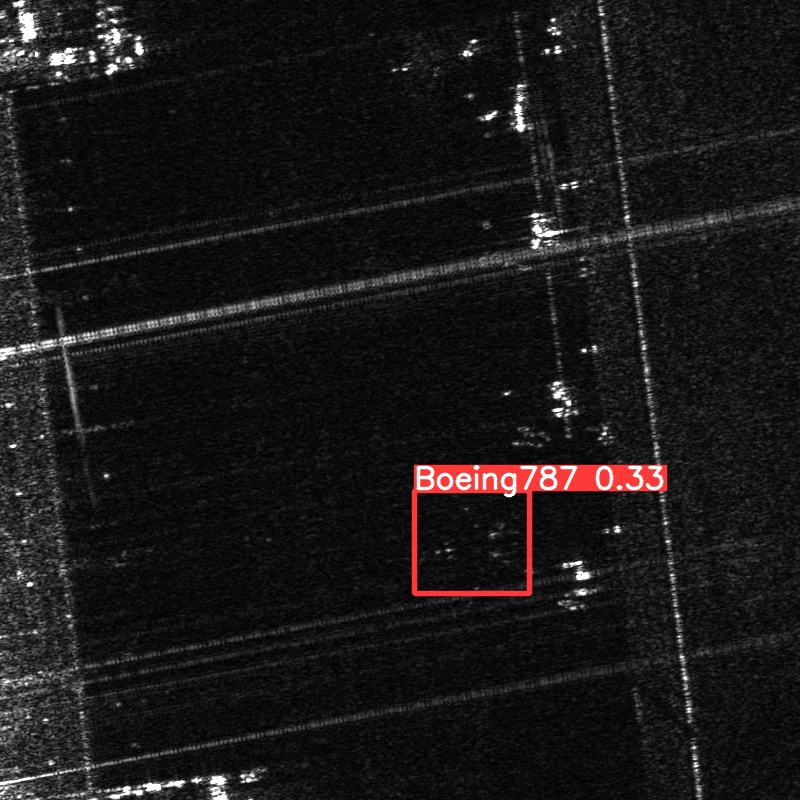}\vskip+3pt
        \includegraphics[width=0.9\linewidth]{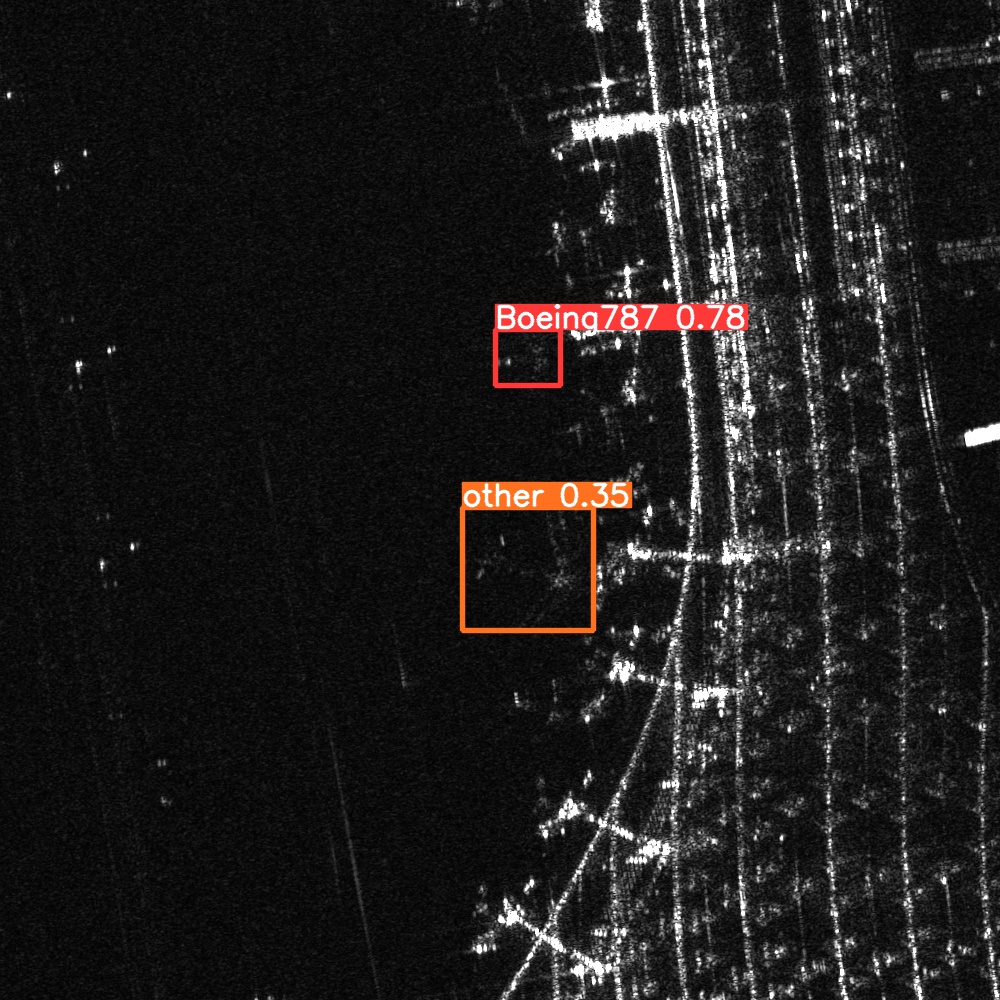}\vskip+3pt
        \includegraphics[width=0.9\linewidth]{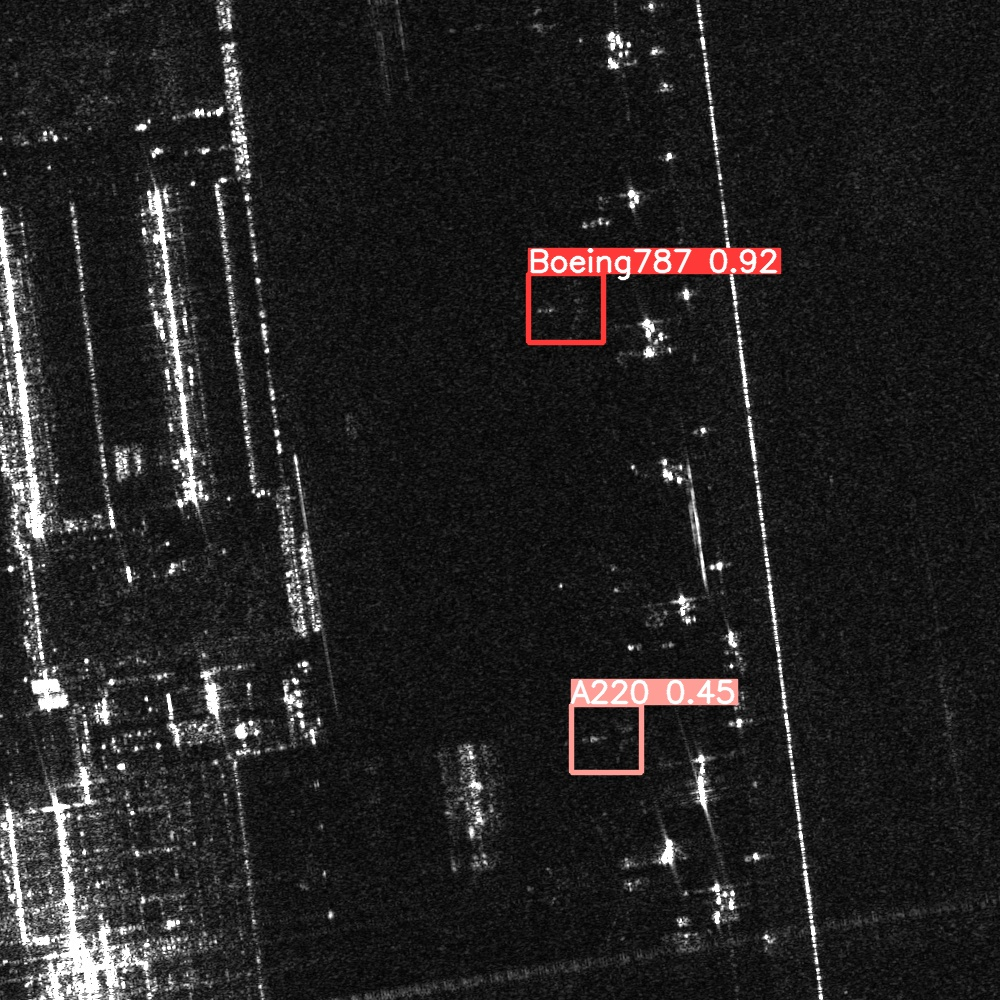}
        \centerline{(c)YOLOv5.}\medskip
    \end{minipage}
    \begin{minipage}[b]{0.19\linewidth} 
      \centering
      \includegraphics[width=0.9\linewidth]{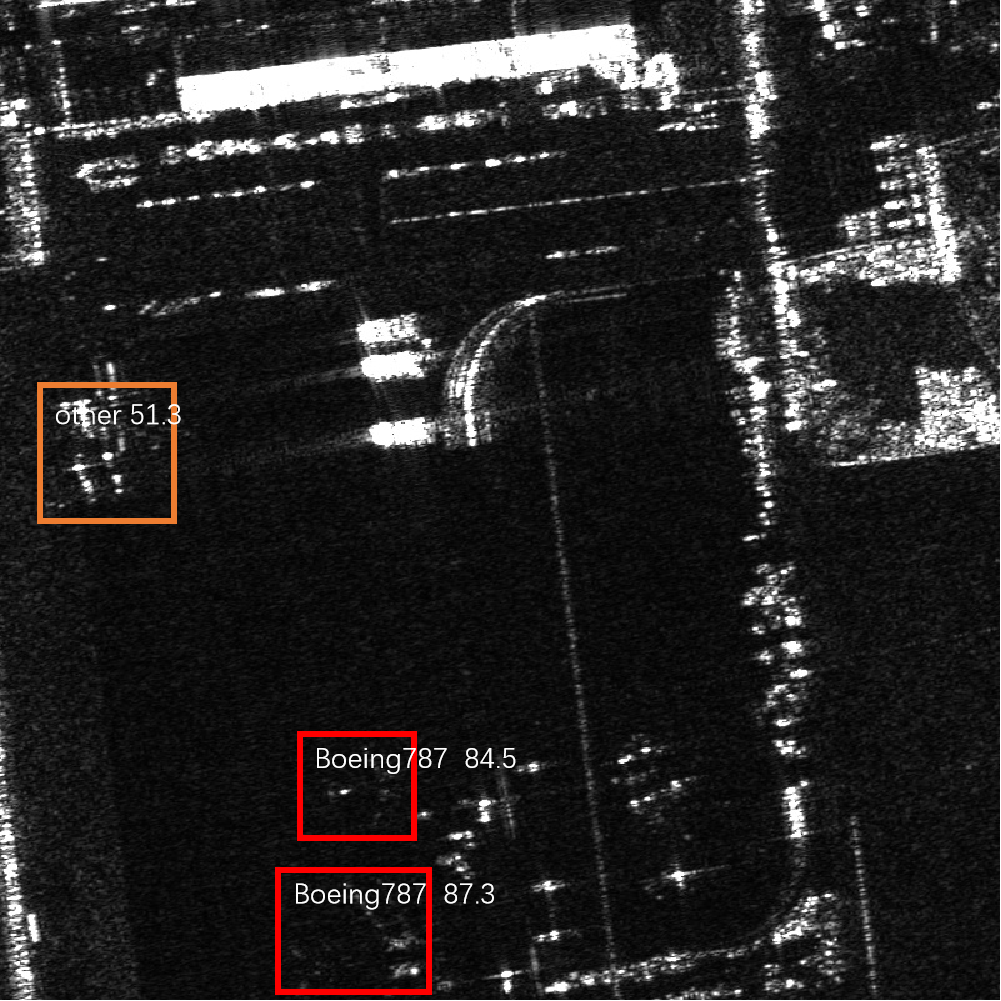}\vskip+3pt
      \includegraphics[width=0.9\linewidth]{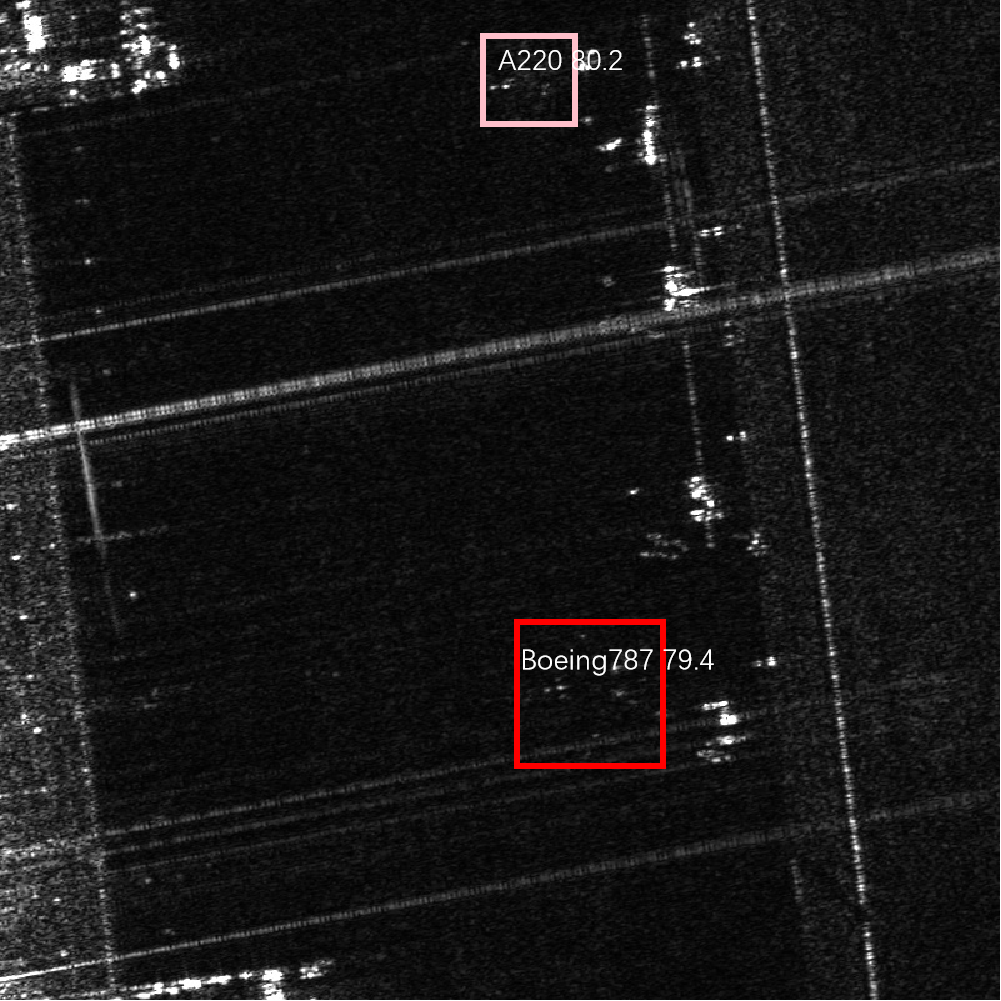}\vskip+3pt
      \includegraphics[width=0.9\linewidth]{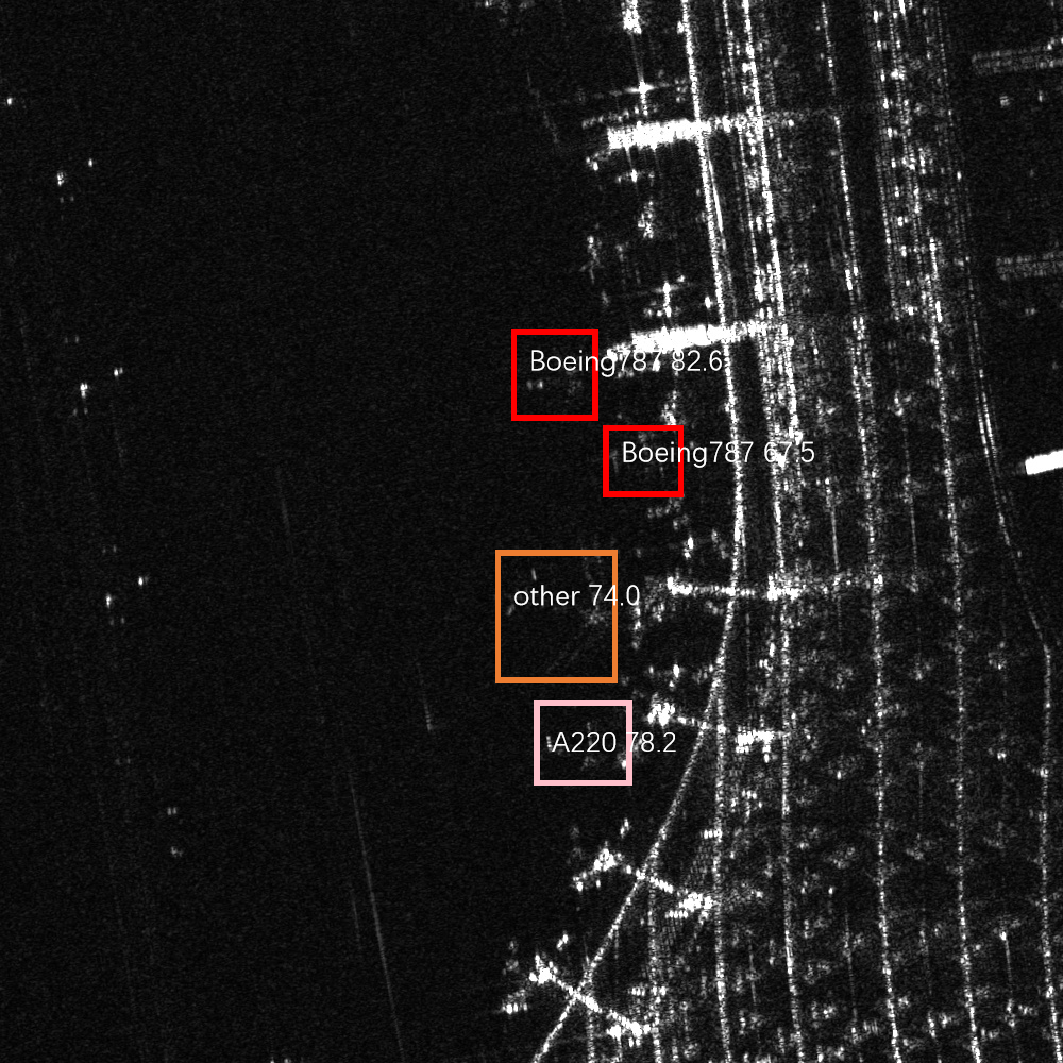}\vskip+3pt
      \includegraphics[width=0.9\linewidth]{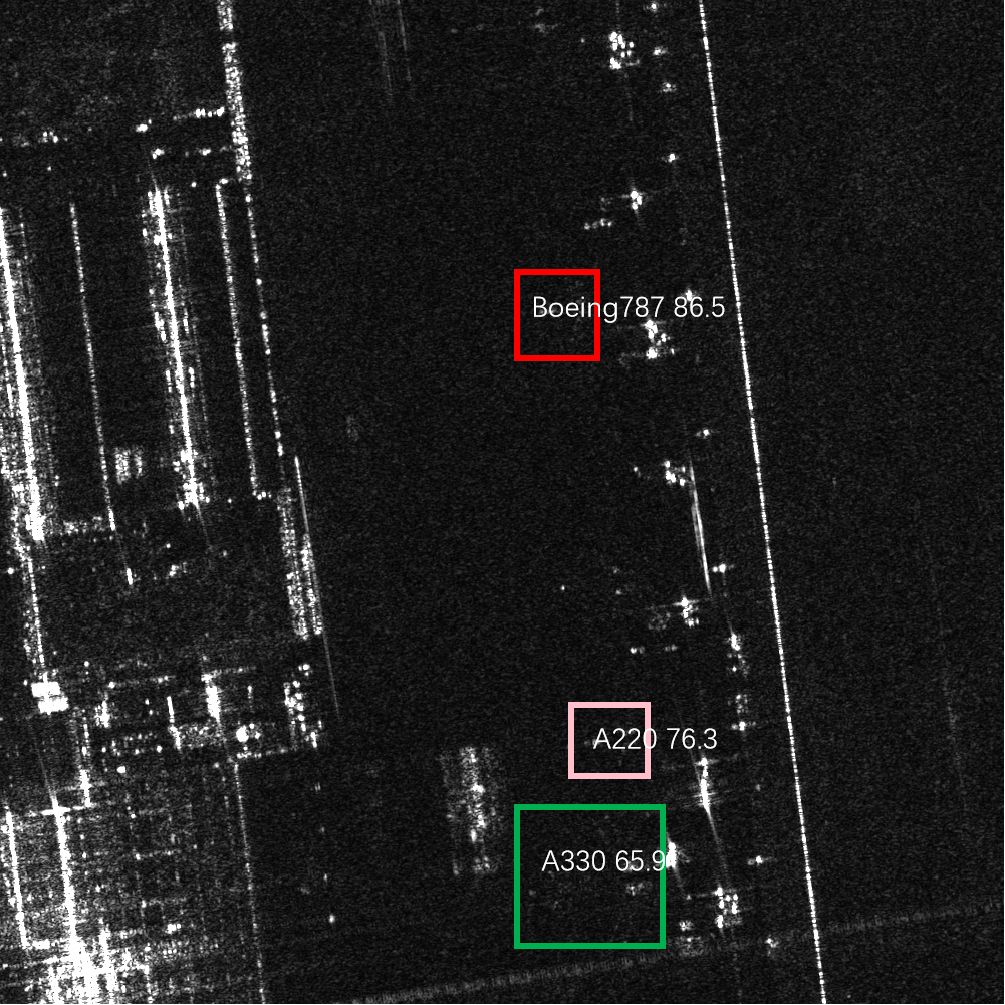}
      \centerline{(d)YOLOv8.}\medskip
    \end{minipage}
    \begin{minipage}[b]{0.19\linewidth} 
        \centering
         \includegraphics[width=0.9\linewidth]{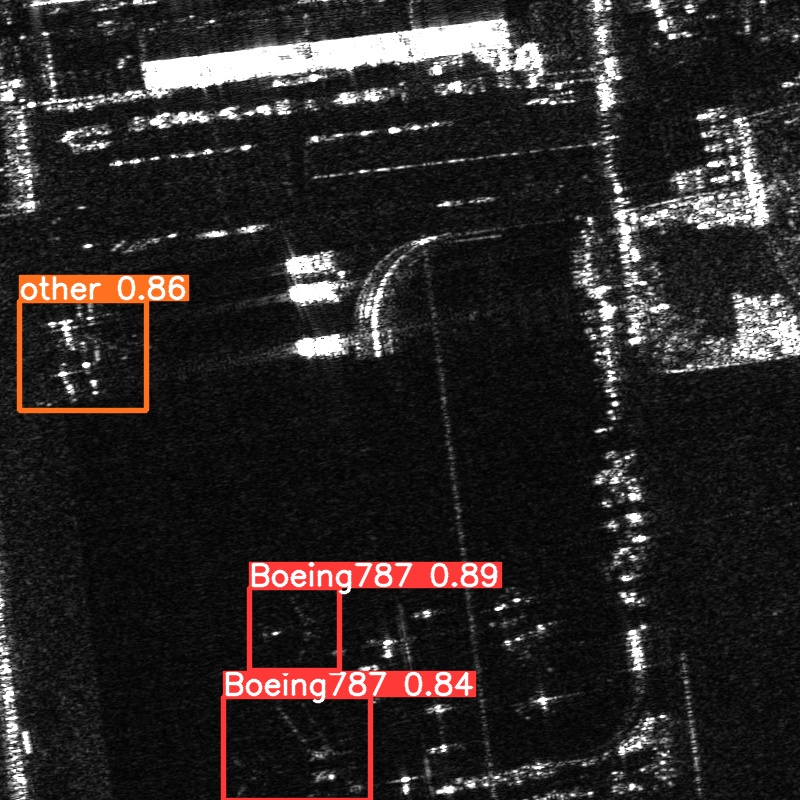}\vskip+3pt
         \includegraphics[width=0.9\linewidth]{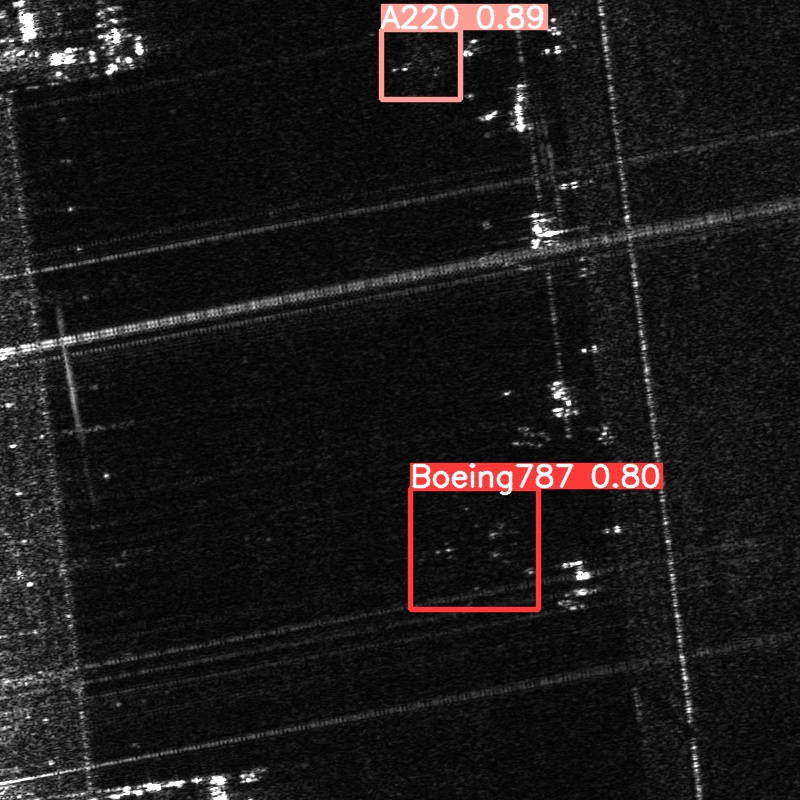}\vskip+3pt
         \includegraphics[width=0.9\linewidth]{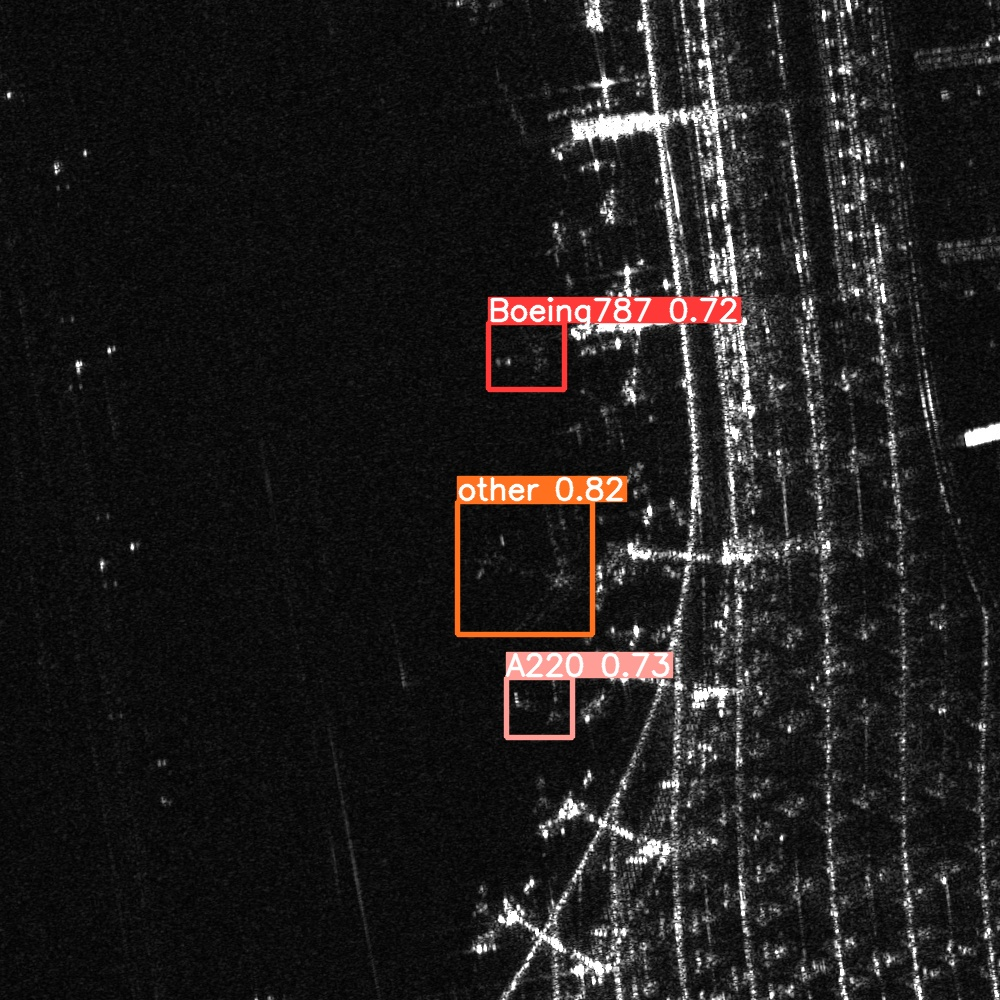}\vskip+3pt
         \includegraphics[width=0.9\linewidth]{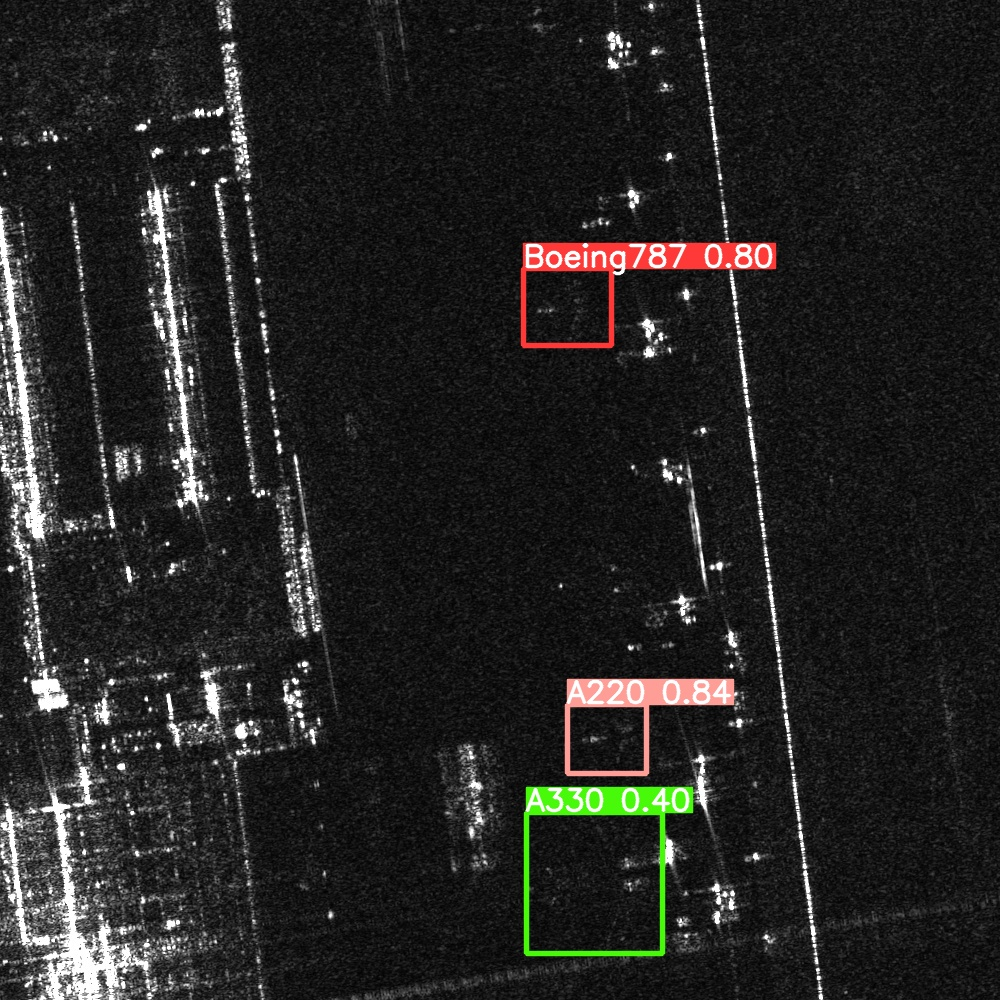}
    \centerline{(e)PGD-YOLO.}\medskip
      \end{minipage}  
      
    \caption{Visualization of the detection results on test data. (a) Ground Truth (b) DEQDet \cite{wang2023deep} (c) YOLOv5 \cite{Jocher_YOLOv5_by_Ultralytics_2020} (d) YOLOv8 \cite{Jocher_Ultralytics_YOLO_2023} (e) PGD-YOLO }
    \label{fig:multi-view}
\end{figure*}

We compare the proposed PGD-YOLO model with 13 object detection methods on SAR-AIRcraft-1.0 dataset, as illustrated in Table \ref{tab:result}. The compared methods include three advanced query-based transformer detectors (DINO \cite{zhang2023dino}, DETReg \cite{bar2022detreg}, DEQDet \cite{wang2023deep}), five one-stage detectors (FCOS\cite{tian2020fcos}, SuperYOLO \cite{zhang2023superyolo}, YOLOv5 \cite{Jocher_YOLOv5_by_Ultralytics_2020}, YOLOv8 \cite{Jocher_Ultralytics_YOLO_2023}, YOLOv10 \cite{wang2024yolov10}), three two-stage detectors (Faster R-CNN\cite{ren2015faster}, Cascade R-CNN \cite{cai2018cascade}, RepPoint \cite{yang2019reppoints}), and two detectors for SAR targets (SKG-Net \cite{fu2021scattering}, SA-Net \cite{zhirui2023sar}). We use the results of FCOS, YOLOv8, Faster R-CNN, Cascade R-CNN, RepPoint, SKGNet and SANet reported in \cite{li2024unleashing}. The others are implemented by us with the same hyper-parameter settings in this paper (300 epoch training). Among them, the query-based transformer detectors generally outperform the one-stage and two-stage detectors. DeTReg \cite{bar2022detreg} obtains an mAP of 87.0\%. DINO can achieve the mAP of 85.2\%. Some popular detection methods, such as Faster R-CNN and Cascade R-CNN, only achieve 76.1\% and 75.7\% mAP, respectively, which are inferior than transformer detectors. As a comparison, our YOLO-based PGD model achieves an mAP of 90.7\%, surpassing other existing mainstream detection methods in computer vision field and SAR community. In addition, our model achieves best or second-best mAP in all categories on the SAR-AIRcraft-1.0 dataset. Specifically, PGD models have remarkable superiority on A220, Boeing737, A330, and ARJ21 categories. The visualization of detection results are depicted in Fig. \ref{fig:multi-view}.

\subsection{Ablation Studies}

In order to assess the soundness of each module in PGD, a sequence of ablation experiments were conducted. 

\subsubsection{Baseline}

The original YOLOv5 employs CSPDarknet and C3 architectures for feature extraction \cite{Jocher_YOLOv5_by_Ultralytics_2020}. It has multiple variants in terms of different depths and widths of backbones, such as YOLOv5-s, YOLOv5-m, YOLOv5-l, YOLOv5-x. We test the detection results as well as the computational cost on these variants, as recorded in Table \ref{table:yolo}. The YOLOv5-x achieves the highest mAP yet being costly. To this end, we optimize the backbone model to alleviate the high cost and reach a good performance based on YOLOv5 architecture. The C3 and CSPDarknet are replaced with a ResNet model with Rep-PAN \cite{li2024yolov}, where the multi-scale features are efficiently extracted and integrated to achieve better representation. As shown in Table \ref{table:yolo}, it achieves an mAP of 83.3\% and has fewer parameters than YOLOv5-m. The modified model is denoted as YOLOv5* and regarded as our baseline model in the following ablation studies.

% By augmenting the depth and receptive field of the network, the network's capacity for feature extraction is enhanced, but at the cost of higher time and parameter count for network inference. For our experiment, we substituted the C3 module and CSPDarknet with Resnet based on Rep-PAN. The RepBlock module is a very efficient feature extraction module that considers the integration of features at various scales. It can improve the network's ability to extract features without dramatically increasing the number of parameters or model complexity. The empirical findings are displayed in Table \ref{table:yolo}. When comparing our yolov5 network to the yolov5m based on the C3 module, we observed a decrease in Params by 2M and an increase in GFLOPs by 30.9. However, we did not compromise on detection accuracy, achieving a mAP of 83.3\%. This shows that our upgraded yolov5 network is more suited for detection tasks.

% \begin{table}[htbp]
% \centering
% \caption{Detection performance of different variants of Yolov5}
% \label{table:yolo}
% \begin{tabular}{c|cccc|c}
% \toprule
% Method  & Recall & Precision & Params & GFLOPs & mAP  \\ \midrule
% YOLOv5s & 76.1   & 77.4      & 7M     & 16.5   & 79.9 \\
% YOLOv5m & 78.8   & 84.4      & 21M    & 48.3   & 83.0 \\
% YOLOv5l & 77.6   & 87.8      & 46M    & 108    & 84.7 \\
% YOLOv5x & 81.0   & 83.0      & 87M    & 204    & 85.5 \\
% YOLOv5* & 70.8   & 86.7      & 19M    & 140.1   & 83.3 \\ 
% \bottomrule
% \end{tabular}
% \end{table}

\begin{table}[htbp]
\centering
\caption{Detection performance of different variants of Yolov5}
\label{table:yolo}
\setlength{\tabcolsep}{5mm}{
\begin{tabular}{ccccc}
\toprule
Models    & mAP   & \#Params(M) \\ 
\midrule
YOLOv5s    & 79.9    & 7       \\
YOLOv5m    & 83.0    & 21       \\
YOLOv5l    & 84.7   & 46      \\
YOLOv5x    & 85.5   & 87         \\
YOLOv5*    & 83.3   & 19    \\ 
\bottomrule
\vspace{-2.0em}
\end{tabular}}
\end{table}

\subsubsection{Ablation Results}

% The proposed PGD mainly contains a self-supervised learning 

The ablation studies presented in Table \ref{tab:ablation} demonstrate the effectiveness of the PGSSL, PGFE, and PGIP modules. In these experiments, PGSSL is implemented using the HRNet architecture, with a simple feature concatenation approach applied to isolate the effect of PGSSL, as shown in the second row of the table. The results indicate that integrating PGSSL into the detector significantly improves mAP by 1.5\%. This enhancement suggests that PGSSL effectively boosts the detector’s capability to correctly identify positive instances. The improvement is attributed to the generalizable feature representations learned by PGSSL, which are highly sensitive to the structural distributions of various SAR airplane targets. A comparison between the second and third rows of Table \ref{tab:ablation} reveals the impact of the PGFE module. Incorporating PGFE, which leverages adaptive feature enhancement guided by physics-aware information from PGSSL, further enhances the detector’s discriminative representation abilities, leading to a 4.1\% increase in mAP. This suggests that PGFE effectively complements PGSSL by refining feature representations based on physics-based guidance. The PGIP module, which is integrated at the detection head, also improves the results by imposing a self-adaptive constraint on dominant scattering structures. It helps the detector better distinguish between true positives and false alarms. By adding the PGIP module, the mAP can be further improved by 1.3\%.

Overall, the proposed PGD framework achieves substantial performance gains, increasing  mAP from 83.3\% to 88.0\%. These results underscore the effectiveness of the integrated modules in enhancing the detection capabilities of the model.

\begin{table}[htbp]
\centering
\caption{Ablation results of different modules.}
\label{tab:ablation}
\begin{tabular}{c|ccc|c}
\toprule
 Baseline                 & PGSSL & PGFE & PGIP & mAP  \\ 
\midrule
\multirow{5}{*}{YOLOv5*} & \ding{55}     & \ding{55}    & \ding{55}    & 83.3 \\ 
% \hline
& \checkmark     & \ding{55}    & \ding{55}    & 84.8 \textcolor{red}{(+1.5)} \\ 
% \hline
& \checkmark     & \checkmark    & \ding{55}    & 87.4 \textcolor{red}{(+4.1)} \\ 
% \hline
 & \ding{55}     & \ding{55}    & \checkmark    & 84.6 \textcolor{red}{(+1.3)} \\ 
% \hline
& \checkmark     & \checkmark    & \checkmark    & 88.0 \textcolor{red}{(+4.7)}\\  \bottomrule
\end{tabular}
\end{table}

\subsection{Discussions on Different Implementations}

\subsubsection{PGSSL Discussion}

\begin{figure*}[htbp]
\centering
\includegraphics[width=1.0\textwidth]{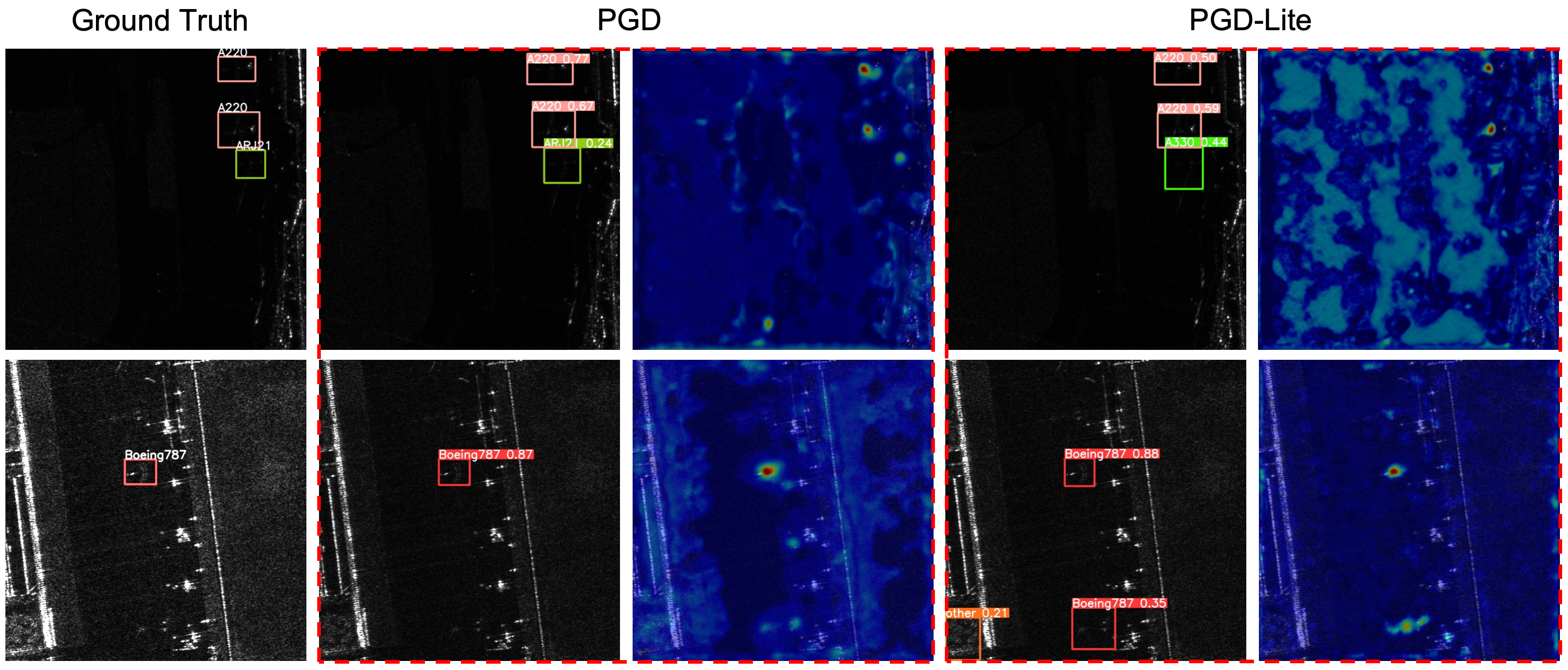}
\caption{The explanation results based on Grad-CAM \cite{selvaraju2017grad} for PGD and PGD-Lite models.}
\label{fig:pgdlite_exp}
\end{figure*}

\begin{figure*}[htbp]
\centering
\includegraphics[width=1.0\textwidth]{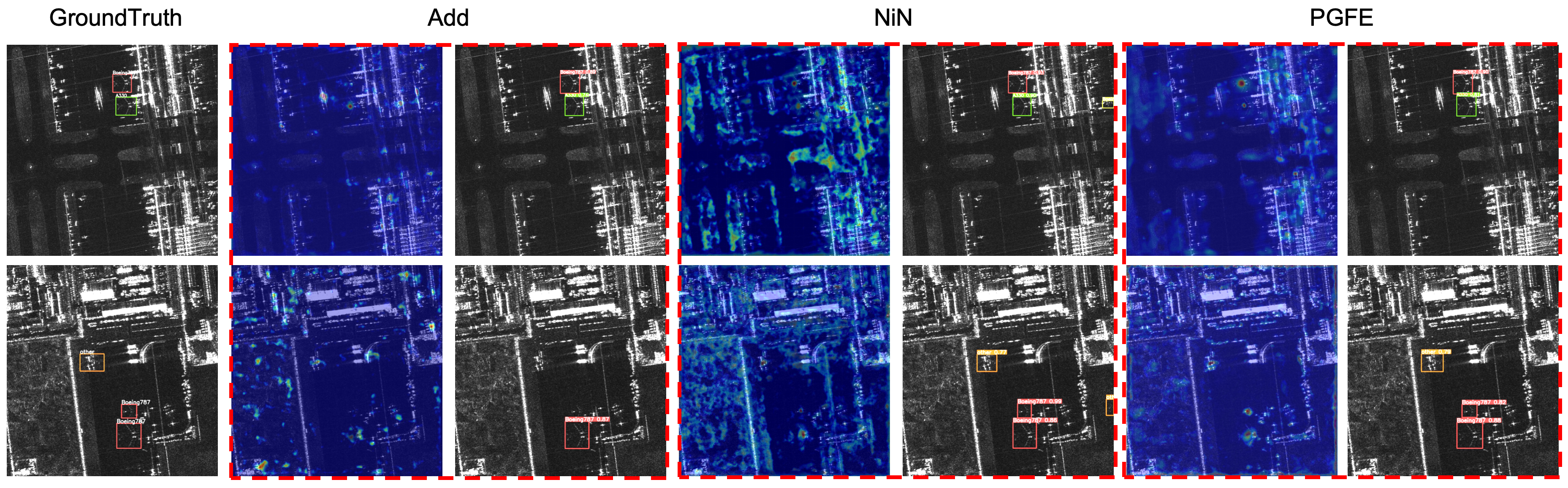}
\caption{The detection results of PGD models that are implemented with different designs of PGFE are presented, as well as the Grad-CAM explanation results of the output feature $F_\mathrm{E}$.}
\label{fig:pgfe_exp}
\end{figure*}

We propose two alternatives for realizing PGSSL, that are, HRNet and TransPose, denoted as PGD and PGD-Lite, respectively. They can both improve the performance of the detector. PGD has more advantage in improving the detection results, while PGD-Lite is more lightweight. The results and computational costs for PGD and PGD-Lite are detailed in Table \ref{tab:skpn}. The results indicate that the HRNet-based PGSSL achieves slightly higher mAP than the TransPose-based implementation. However, the TransPose-based approach requires less computational resources, making it more efficient in terms of time and resource usage. Given these factors, while the HRNet-based PGSSL offers a marginal performance advantage, the TransPose-based PGSSL is a more time-efficient and resource-efficient option.

% TransPose applied a Transformer-based architecture to extract the multi-scale features from the input image 

% \begin{table}[htbp]
% \centering
% \caption{Comparison of different implementations for physics-guided self-supervised learning (PGSSL).}
% \label{tab:skpn}
% \begin{tabular}{cccccc}
% \toprule
% PGSSL    & #Params(M)  & GFLOPs   & Recall & Precision & mAP           \\ 
% \midrule
% Base     & 19      & 140.1        & 70.8   & 86.7     & 83.3 \\
% HRNet     & 48(+29)      & 271.6        & 75.5(+4.7)   & 85.7(-1.0)      & 84.8(+1.5) \\ 
% % \hline
% TransPose & 28(+9) & 236.0 & 74.8(+4.0)   & 85.8(-0.9)      & 83.9(+0.6)          \\ \bottomrule
% \end{tabular}
% \end{table}

\begin{table}[htbp]
\centering
\caption{Comparison of different implementations for physics-guided self-supervised learning (PGSSL).}
\label{tab:skpn}
\begin{tabular}{cccc}
\toprule
PGSSL    & \#Params(M) &FPS   & mAP           \\ 
\midrule
Baseline     & 19         &33      & 83.3 \\
PGD     & 48     &8        & 84.8 \\ 
% \hline
PGD-Lite & 28     &18   & 83.9          \\ \bottomrule
\end{tabular}
\end{table}

% Additionally, we apply Grad-CAM \cite{selvaraju2017grad} approach for further explanation. Fig. \ref{fig:pgdlite_exp} visualizes the attention maps of features $F_\mathrm{E}$ in PGD and PGD-Lite models, respectively. The detection results of PGD and PGD-Lite are given in (b) and (c), where the confidence scores of the two A220 predictions by PGD-Lite are lower than those predicted by PGD. According to the attention maps generated with Grad-CAM explanation method given in (d) and (e), PGD and PGD-Lite can successfully concern about the dominant scattering information of SAR airplane targets and PGD model is able to learn more discriminative features.

Additionally, we applied the Grad-CAM approach \cite{selvaraju2017grad} for further interpretability analysis of our models. Fig. \ref{fig:pgdlite_exp} visualizes the saliency maps of the feature representations $F_\mathrm{E}$ for both the PGD and PGD-Lite models. The detection results, shown in (b) and (c), indicate that both PGD and PGD-Lite can achieve a good detection performance, but the confidence scores for the two A220 predictions made by PGD-Lite are lower than those made by the PGD model. The Grad-CAM-generated attention maps, presented in (d) and (e), reveal that both PGD and PGD-Lite effectively focus on the dominant scattering characteristics of the SAR airplane targets. However, the PGD model demonstrates an enhanced ability to learn more discriminative features, as evidenced by its more focused and precise saliency maps.

% \begin{figure*}[htbp]
% \centering
% \includegraphics[width=1.0\textwidth]{fig_pgdlite_exp.png}
% \caption{The explanation results based on Grad-CAM \cite{selvaraju2017grad} for PGD and PGD-Lite models.}
% \label{fig:pgdlite_exp}
% \end{figure*}

% \begin{figure*}[htbp]
% \centering
% \includegraphics[width=1.0\textwidth]{fig_pgfe_exp.png}
% \caption{The detection results of PGD models that are implemented with different designs of PGFE are presented, as well as the Grad-CAM explanation results of the output feature $F_\mathrm{E}$.}
% \label{fig:pgfe_exp}
% \end{figure*}

\subsubsection{PGFE Discussion}

\begin{figure*}[!htbp]
\centering
\includegraphics[width=0.9\textwidth]{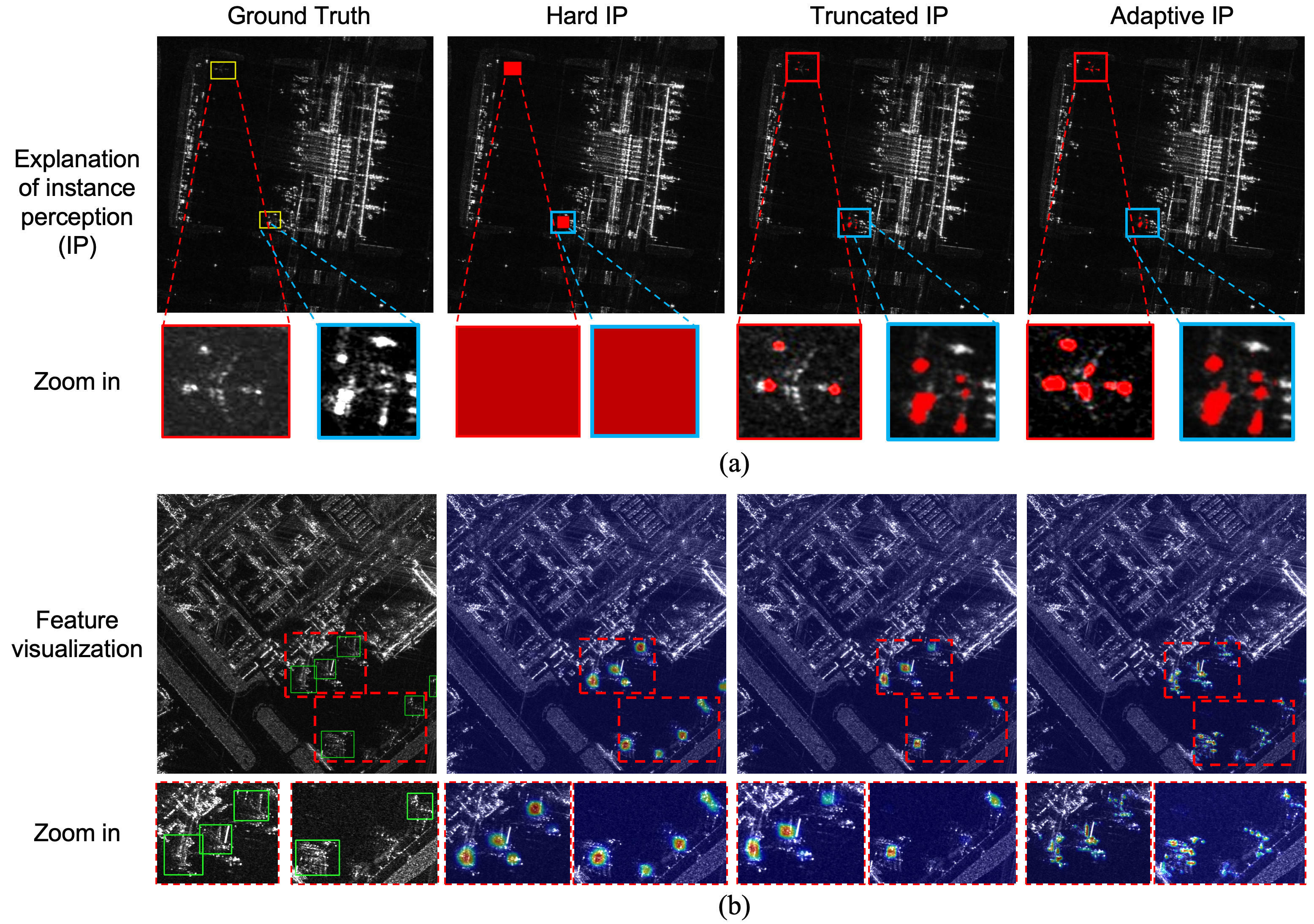}
\caption{(a) The illustration of the hard and truncated instance perception (IP) methods, compared with the proposed adaptive IP. (b) The visualization of the learned features at the detection head, trained with hard IP, truncated IP, and adaptive IP.}
\label{fig:IPexp}
\end{figure*}

The PGFE module is designed to enhance the feature representation of targets for detection by leveraging physics-aware embeddings learned from PGSSL. We experimented with different feature fusion strategies to demonstrate the effectiveness of PGFE, as detailed in Table \ref{tab:caf}.

\begin{table}[htbp]
\centering
\caption{Comparison of different implementations for physics-guided feature enhancement (PGFE).}
\label{tab:caf}
\begin{tabular}{cccc}
\toprule
PGFE & \#Params(M) & GFLOPS & mAP           \\ \midrule
AddFusion &48    &271.2       & 84.8          \\ 
NINFusion &49    &272.9       & 85.9          \\ 
Ours &50    &279.0       & 87.4 \\ \bottomrule
\end{tabular}
\end{table}

Initially, we applied the AddFusion strategy, where multi-scale features from the Feature Pyramid Network (FPN) are directly added to the pre-trained physics-aware features from PGSSL. This approach does not introduce additional parameters and achieves a mean Average Precision (mAP) of 84.8\%. Next, we evaluated the NINFusion \cite{DBLP:conf/bmvc/Li0T018}, which concatenates features and processes them using a Network-in-Network architecture \cite{DBLP:journals/corr/LinCY13}. Compared to AddFusion, NINFusion improves the mAP by 1.1\%, indicating its effectiveness in better integrating features.

% \begin{figure*}[htbp]
% \centering
% \includegraphics[width=1.0\textwidth]{fig_pgfe_exp.png}
% \caption{The detection results of PGD models that are implemented with different designs of PGFE are presented, as well as the Grad-CAM explanation results of the output feature $F_\mathrm{E}$.}
% \label{fig:pgfe_exp}
% \end{figure*}

% \begin{figure*}[!htbp]
% \centering
% \includegraphics[width=0.9\textwidth]{fig_IPexp.png}
% \caption{(a) The illustration of the hard and truncated instance perception (IP) methods, compared with the proposed adaptive IP. (b) The visualization of the learned features at the detection head, trained with hard IP, truncated IP, and adaptive IP.}
% \label{fig:IPexp}
% \end{figure*}

However, both of these strategies treat the FPN features and physics-aware (PA) features equally, overlooking their distinct contributions to detection. The PA features from PGSSL are adept at capturing various SAR airplane representations but are less effective at characterizing other targets. In contrast, FPN features from the original detector are tailored to represent complex scenes with diverse scattering patterns. Therefore, a more nuanced approach that considers the complementary nature of these features is necessary.

The proposed PGFE employs a cross-attention-based module, where PA features are treated as the query, and FPN features serve as both the value and key. This mechanism re-weights the multi-scale FPN features according to the physics-aware information provided by PGSSL, resulting in features that are more discriminative for target representation while maintaining robust representation for complex scenes. The experimental results demonstrate that the proposed PGFE achieves the highest mAP of 87.4\% and the params only increased by 2M, outperforming other fusion strategies.

Additionally, we utilize Grad-CAM to provide a deeper analysis of the various fusion strategies employed for integrating the physics-aware features from PGSSL with the detection model’s features. The corresponding saliency maps are presented in Fig. \ref{fig:pgfe_exp}. These maps highlight the key discriminative features used for target detection. The results reveal that both the simple addition and NIN fusion strategies fail to fully capitalize on the advantages of the pre-trained PA features. These methods do not effectively complement the unique representations provided by PGSSL and the original detection model. In contrast, the proposed PGFE method shows a marked improvement in focusing on the airplane targets, significantly reducing the interference from the surrounding complex scene. This demonstrates the superior capability of PGFE in leveraging physics-aware information to enhance target detection.

\subsubsection{PGIP Discussion}
% \textbf{Evaluation of different corner extraction methods.}

% \begin{figure*}[!htbp]
% \centering
% \includegraphics[width=0.9\textwidth]{fig_IPexp.png}
% \caption{(a) The illustration of the hard and truncated instance perception (IP) methods, compared with the proposed adaptive IP. (b) The visualization of the learned features at the detection head, trained with hard IP, truncated IP, and adaptive IP.}
% \label{fig:IPexp}
% \end{figure*}

In the PGIP module at the detection head, we introduce an adaptive instance perception (IP) learning strategy to emphasize the refined, discrete features of SAR airplane targets. We compare this adaptive IP method with two alternative strategies: hard IP and truncated IP, as illustrated in Fig. \ref{fig:IPexp}. In the hard IP approach, semantic supervision  $L_i$  is applied uniformly across a rectangular region defined by the annotated bounding box, with all pixels within this box assigned a value of 1 for loss calculation. Conversely, the truncated IP strategy focuses on supervising only the strong scattering points within the image, a technique frequently used in related literature. This method employs a global parameter to filter out strong scattering pixels, but it does not account for the relative variations in backscattering that different SAR airplanes may exhibit.
\begin{table}[htbp]
\centering
\caption{Comparison of different implementations for physics-guided instance perception (PGIP).}
\label{tab:SRL}
\begin{tabular}{cccc}
\toprule
PGIP            & Hard IP & Truncated IP & Truncated IP \\ \midrule
mAP      & 83.9   & 84.2      & 84.6         \\ \bottomrule
\end{tabular}
\end{table}

The visualized results indicate that the hard IP strategy captures only the coarse features of each target, neglecting the specific discrete characteristics that are crucial for accurate detection. Although the truncated IP method can further refine these discrete features, its reliance on a global truncation threshold results in the omission of some important features with relatively weaker backscattering, such as the wings that exhibit primary surface scattering characteristics. In contrast, the proposed adaptive IP approach effectively captures both the discrete and critical local features of SAR airplanes, leading to improved detection performance, as demonstrated in Table \ref{tab:SRL}.

\section{Conclusion}
\label{sec:conclusion}

In this paper, a novel physics-guided detector (PGD) learning paradigm is proposed for SAR airplanes and it can be extended to various deep learning-based detectors to facilitate the generalization ability and alleviate overfitting. The motivation is to comprehensively leverage the implicit prior knowledge in the discrete and variable structure distributions of a wide range of SAR airplanes obtained from multiple sources. The proposed PGD has three main contributions, including PGSSL, PGFE, and PGIP. PGSSL is constructed as a distribution heatmap prediction task by introducing some external knowledge of various airplane images, which aims to learn the efficient and generalizable representations aware of the structure distribution of SAR airplanes. Subsequently, PGFE is designed as a cross-attention-based fusion module to guide the backbone features of the detector to pay more attention to the discrete structure distributions of targets. PGIP at the detection head constrains the detector from focusing the dominate scattering points rather than the complex background. We demonstrate the effectiveness of the proposed PGD with extensive experiments on SAR-AIRcraft-1.0 dataset, where PGD is applied to different deep learning-based detectors for evaluation. We discussed several implementations of PGD, demonstrating the flexibility and effectiveness of the proposed method.

\bibliography{IEEEabrv,ref}
% You can manually copy in the resultant .bbl file and set second argument of $\backslash${\tt{begin}} to the number of references
%  (used to reserve space for the reference number labels box).

\bibliographystyle{IEEEtran}

\vfill

\end{document}